\documentclass{article}

\usepackage[preprint]{neurips_2023}

\usepackage[utf8]{inputenc} 
\usepackage[T1]{fontenc}    
\usepackage{hyperref}       
\usepackage{url}            
\usepackage{amsfonts}       
\usepackage{nicefrac}       
\usepackage{microtype}      
\usepackage{xcolor}         
\usepackage{inconsolata}
\usepackage{amssymb}   
\usepackage{graphicx}  
\usepackage{amsmath}
\usepackage{subfigure} 
\usepackage{makecell} 
\usepackage{enumitem} 
\usepackage{multirow} 
\usepackage{booktabs} 
\usepackage{stfloats}
\usepackage{wrapfig}
\usepackage{cite}
\usepackage{color} 
\usepackage{algorithm}
\usepackage{algorithmic}

\usepackage{newfloat}
\usepackage{listings}
\floatstyle{ruled}
\newfloat{listing}{tb}{lst}{}
\floatname{listing}{Listing}

\usepackage{graphicx}
\usepackage{amsmath,amssymb,amsthm,mathabx,amsfonts}
\usepackage{bbm}
\usepackage{acronym}
\usepackage{enumitem}
\usepackage{cleveref}
\usepackage{balance}
\usepackage{xspace}
\usepackage{setspace}
\usepackage[skip=3pt]{caption}
\usepackage{url}
\usepackage{amsfonts}
\usepackage{multirow}
\usepackage{booktabs}
\usepackage{tablefootnote}
\usepackage{multicol,lipsum}
\usepackage{tikz}
\usepackage{pgfplots}
\usepackage{pgfplotstable}
\usepackage{booktabs}

\usepackage{arydshln}

\usepackage{colortbl}
\definecolor{gblue}{HTML}{4285F4}
\definecolor{gred}{HTML}{DB4437}
\definecolor{ggreen}{HTML}{0F9D58}

\definecolor{mygray}{gray}{.92}
\definecolor{emphypurple}{rgb}{0.302, 0.055, 0.659}

\definecolor{highlightgreen}{HTML}{009901}
\definecolor{highlightred}{HTML}{FD6864}

\pgfplotsset{compat=1.18}
\makeatletter
\DeclareRobustCommand\onedot{\futurelet\@let@token\@onedot}
\def\@onedot{\ifx\@let@token.\else.\null\fi\xspace}
 
\def\eg{\emph{e.g}\onedot} 

\def\ie{\emph{i.e}\onedot}

\def\etc{\emph{etc}\onedot}

\makeatother

\makeatletter
\newcommand{\thickhline}{%
    \noalign {\ifnum 0=`}\fi \hrule height 1pt
    \futurelet \reserved@a \@xhline
}

\DeclareMathOperator*{\argmax}{arg\,max}

\acrodef{nlp}[NLP]{natural language processing}
\acrodef{llm}[LLM]{Large Language Model}
\acrodef{sota}[SOTA]{state-of-the-art}
\acrodef{bs}[BS]{Beam Search}
\acrodef{mhs}[MHS]{Metropolis-Hastings Sampling}
\acrodef{hs}[HS]{Hybrid Search}
\acrodef{uas}[UAS]{unlabeled attachment score}
\acrodef{dda}[DDA]{Directed Dependency Accuracy}
\acrodef{sota}[SOTA]{state-of-the-art}
\acrodef{pos}[POS]{part-of-speech}
\acrodef{asr}[ASR]{attacking success rate}
\acrodef{ppl}[PPL]{Perplexity score}
\acrodef{mcmc}[MCMC]{Markov Chain Monte Carlo}
\acrodef{dp}[DP]{dependency parsing}

\acrodef{bbl}[BBL]{BIG-bench Lite}
\acrodef{plm}[PLM]{Pre-trained Language Model}
\acrodef{p2p}[\texttt{P2P}]{\texttt{Prompt to Prompt}}

\crefname{algorithm}{Alg.}{Algs.}
\Crefname{algocf}{Algorithm}{Algorithms}
\crefname{section}{Sec.}{Secs.}
\Crefname{section}{Section}{Sections}
\crefname{table}{Tab.}{Tabs.}
\Crefname{table}{Table}{Tables}
\crefname{figure}{Fig.}{Fig.}
\Crefname{figure}{Figure}{Figure}
\crefname{appendix}{Appendix}{Appendices}
\usepackage{natbib}
\usepackage{xcolor}
\usepackage{multirow}
\usepackage{array}

\newcommand{\model}[0]{MCS~}
\newcommand{\cail}[0]{CAMLOP~}

\title{Human-in-the-Loop through Chain-of-Thought}

\author{%
Zefan Cai$^{1, 2}$,
\textbf{Baobao Chang}$^{1}$\footnotemark[1],
\textbf{Wenjuan Han}$^{3}$\footnotemark[1],
\\
$^1$National Key Laboratory for Multimedia Information Processing, Peking University \\ 
$^2$School of Software and Microelectronics, Peking University, China \\
$^3$Beijing Jiaotong University, Beijing, China \\
\texttt{zefncai@gmail.com}; \texttt{chbb@pku.edu.cn}; \texttt{wjhan@bjtu.edu.cn}; \\
}

\begin{document}

\maketitle

\renewcommand{\thefootnote}{\fnsymbol{footnote}}
\footnotetext[1]{Corresponding authors.}
\renewcommand{\thefootnote}{\arabic{footnote}}

\begin{abstract}
While the emergence of powerful language models along with Chain-of-thought prompting has made automation more and more omnipresent, it sometimes demonstrates its weakness in long-term or multi-step logical reasoning. For example, users don't always get desirable answers for complex mathematical problems without human involvement. Against this background, we present the \textbf{Manual Correction System (MCS)} --- a human-in-the-loop system enhanced by Chain-of-Thought prompting, which explores how manual correction of sub-logics in rationales can improve LLM's reasoning performance. Moving one step forward, considering a system with human-in-the-loop involves more than having humans improve performance but also controlling the cost. Therefore, we post a \textbf{Cost-utility Analysis Model for Human-in-the-Loop systems (CAMLOP)} based on classical economics theory to analyze, quantify and balance the utility and the corresponding cost. We conduct experiments of MCS and CAMLOP with twelve datasets. A significant advantage w.r.t cost and utility proves its superiority over strong baselines.
\end{abstract}
\section{Introduction}

Large language model-based Artificial Intelligence  systems are augmenting humans in certain roles, and soon this trend will expand to the vast majority of the workforce. However, while the emergence of powerful language models \citep{sanh2021multitask,ouyang2022training, zhang2022opt, shao2023compositional} has made automation omnipresent, it sometimes demonstrates its weakness in long-term or multi-step logical reasoning \citep{hosseini2014learning,kushman2014learning, koncel2015parsing,roy2016solving}. For example, users don't always get desirable answers for a mathematical problem without human involvement. 
To make tangible progress in mitigating these errors is where we need humans, and a system with human-in-the-loop involves more than having humans improve performance but also controlling the cost.
Against this background, there comes a timing question: how to get a human-in-the-loop system in the most effective (namely, high-utility) and low-cost way?

\begin{figure*}[ht]
\centering
\includegraphics[width=1.0\linewidth]{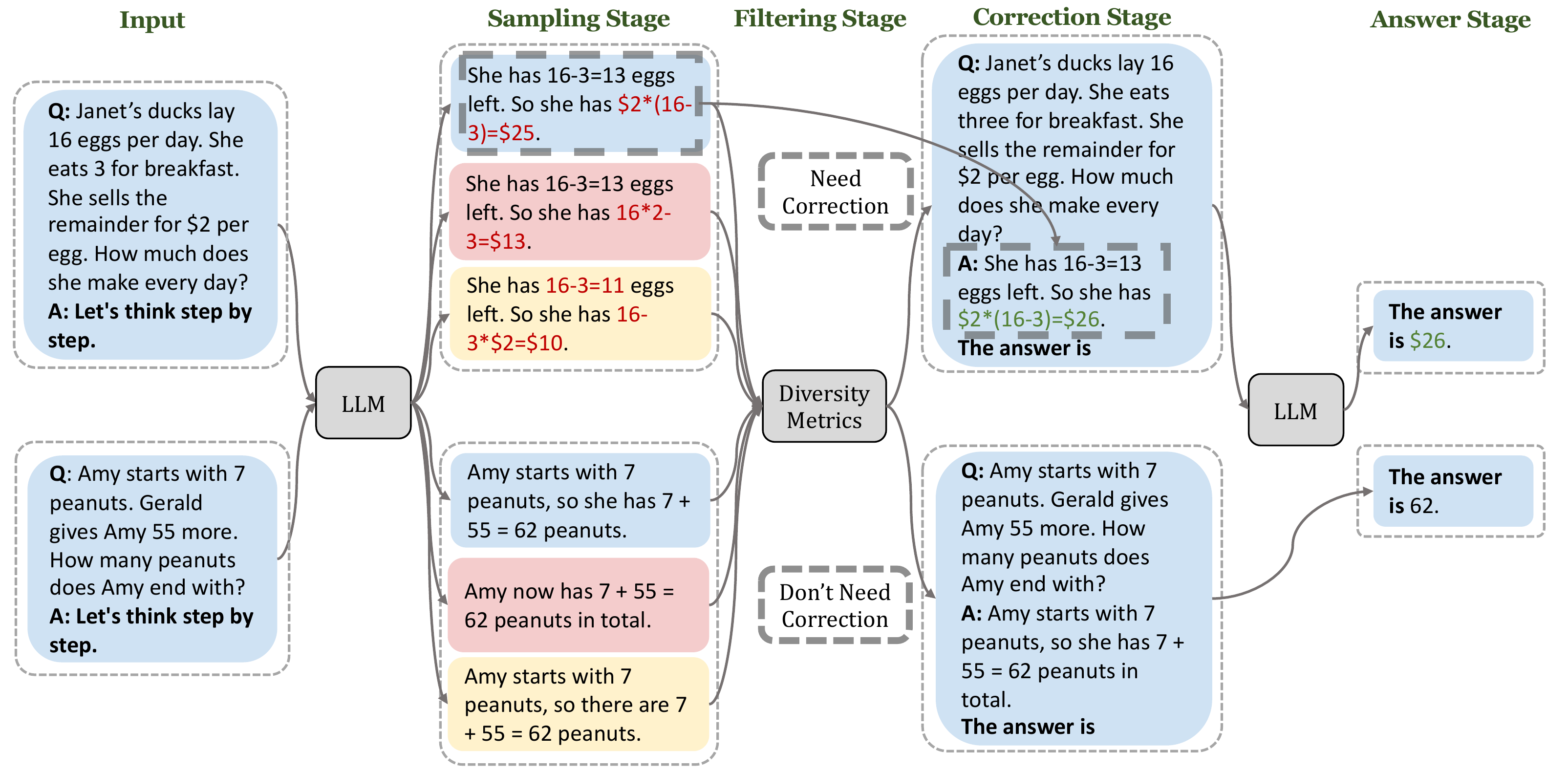}
\caption{
\model comprises four stages: (1) \textbf{sampling stage} prompting the LLM using CoT prompting and replacing the greedy decoding by sampling from the LLM’s decoder to generate a set of rationales (\ie, the complete logical chain of CoT output); (2) \textbf{filtering stage} filtering out the samples ranked high by Diversity Entropy; (3) \textbf{correction stage} manually adding, deleting and modifying erroneous sub-logics in the most likely rationale of the filtered sample, and (4) \textbf{answer stage} prompting the LLM using CoT prompting again with manually corrected sub-logics and using greedy decoding to obtain the final answer.
}
\label{fig:illustrationmodel}
\end{figure*}

See Fig.~\ref{fig:illustrationmodel} as an example. For humans, solving the whole problem in the leftmost box is often more difficult than solving one of the sub-logics (\eg, $2*(16-3) =25)$. Correction of the erroneous sub-logic (\eg, $2*(16-3) =25 \rightarrow 2*(16-3) =26$) helps LLM reach a correct final answer. 

In the last few years, thanks to explorations in Large Language Models (LLMs) and advances in in-context learning (ICL) technologies, giant breakthroughs have been obtained. Just by being fed an instruction, models can function very well on that task without manual finetuning~\citep{NEURIPS20201457c0d6}. This provides a chance for a human to change the predicted results via natural language instructions as a flexible and friendly interface. Furthermore, changing the rationale for chain-of-thought (CoT) prompting~\citep{wei2022chain} is even more user-friendly since short and simple sub-logics in the rationale are easy for humans to handle.
Whereas manual correction helps, the labor of this additional correction stage brings a direct and indirect cost (See Sec.~\ref{sec:framework} for more details). When 
and how humans intervene will greatly affect the cost and utility. Until recently, few researchers had explored this balance in ICL.

We present the \textbf{M}anual \textbf{C}orrection \textbf{S}ystem (\model; Sec.~\ref{sec:methodology}) --- a human-in-the-loop system, which explores when and how manual correction of rationales can efficiently improve LLM's reasoning ability. To our knowledge, \model is the first human-in-the-loop system leveraging rationales. As shown in Fig. \ref{fig:illustrationmodel}, \model consists of four stages: prompting the LLM with CoT, automatically filtering out the incorrectly predicted samples, human correcting their rationales, and prompting the LLM using CoT again to obtain the final answer. 
Referring to the ``when'' problem, we consider a diversity-based method to get a cue to indicate when humans should be involved, so as to reduce human labor as much as possible (See.~\ref{subsection:filtering}). The diversity-based method is inspired by the diversity of the rationales.
We have found that even when the desired answer is fixed, introducing the diversity degree of the rationales can be highly beneficial; therefore we introduce Diversity Metrics, as commonly used in Active Learning field \citep{brinker2003incorporating,yang2015multi,agarwal2020contextual}, to find data points requiring manual intervention. Then it comes to the ``how'' problem (See.~\ref{subsection:manual_correction}). We empirically prove the viability of paying attention to sub-logics instead of the whole problem. We define three operations (\ie, modifying, adding, and deleting) that a human can perform on the sub-logics of rationales for efficiency and simplification.

With the development of Artificial Intelligence (AI), some companies have started to explore the use of LLMs in practice (\eg, IBM implementing AI processes in HR~\citep{IBM}).
Therefore, we propose a \textbf{C}ost-utility \textbf{A}nalysis \textbf{M}odel for Human-in-the-\textbf{LO}o\textbf{P} systems (\cail; Sec.~\ref{sec:framework}) to analyze and balance the cost and utility. \cail describes the cost-utility ratio that is introduced from the economics theory into the AI field to quantify these two factors (\ie, cost and utility) and spread the two factors across various aspects (\eg, time and money as cost; accuracy and user satisfaction as utility) so that reliable scores of various aspects are achieved.

We instantiate \model with  twelve datasets across three classes of tasks --- arithmetic, commonsense, and symbolic reasoning (Sec.~\ref{sec:experiments}).
\model achieves new state-of-the-art levels of performance across most of the tasks. 
To show the applicability in real-world business, we apply \cail to practice by posing an example to illustrate the balance between utility and cost in Sec.~\ref{sec:analysis_cost_utility}. Notably, a significant advantage w.r.t cost and utility proves our \model's superior over strong baselines. 

\section{Manual Correction System}
\label{sec:methodology}


MCS automatically finds the incorrectly predicted samples to indicate when humans should be involved (Sec. \ref{subsection:filtering}) and then provides efficient operations to indicate how to correct rationales (Sec. \ref{subsection:manual_correction}). Fig.~\ref{fig:illustrationmodel} shows the whole four stages in MCS. The first and final stages are simple prompting. The intermediate filtering stage and correction stage are our focus, as detailed below.

\subsection{Filtering Stage}
\label{subsection:filtering}
As shown in Fig.~\ref{fig:illustrationmodel}, after the first stage, the LLM samples three plausible rationales for a math problem that arrive at different answers. Just like humans, LLMs may make countless and various mistakes, but there are only a limited number of correct rationales for the right result. If most of the sampled rationales cannot make agreements, with a high probability this sample is wrongly predicted.
To empirically prove that, we conduct quantitative experiments and discover that incorrectly predicted samples tend to have greater diversity in their final answer when solving difficult reasoning problems. (Please refer to Appendix ~\ref{appendix:experiments_for_filtering_stage} for more details).

Specifically, the LLM is prompted with a set of manually written CoT exemplars following ~\citet{wei2022chain} in the first stage.  (Please refer to Appendix for more details) Then, we sample a set of candidate outputs from the LLM's decoder to generate a set of rationales\footnote{Most existing sampling algorithms including temperature sampling~\citep{ackley1985learning, ficler2017controlling}, top-$k$ sampling~\citep{fan2018hierarchical, holtzman2018learning, radford2019language} and nucleus sampling~\citep{holtzman2019curious} could be used for sampling the required rationals. Here we follow \citet{wang2022self} for a fair comparison. Other sampling methods can also bring a general benefit.}. Finally, we use the diversity degree to identify the most likely incorrect sample for humans to involve. Here, we adopt a widely-used method to select the samples: Diversity Entropy~\citep{brinker2003incorporating,yang2015multi,agarwal2020contextual}. A further study about Diversity Entropy in Sec.~\ref{paragraph:diversity_metrics_over_diverse_reasoning_paths} quantitatively demonstrates its advantage.

Formally, given a manually written CoT prompt and a sample $\mathbf{s}$, \model decodes a set of $N$ outputs, where each output $\mathbf{r}_i$ is a sequence of tokens representing the $i$-th rational, then the rational $\mathbf{r}_i$ is used to obtain the answer $\mathbf{a}_i$. 
As previously demonstrated, a greater diversity of the set of answers indicates potential incorrect predictions and flags a sample for humans to involve. First, we obtain the predicted answer $\mathbf{a}_i$ though $\argmax_{\mathbf{a}_i} P(\mathbf{r}_i, \mathbf{a}_i \mid \mathbf{s})$. For example, in Fig.~\ref{fig:illustrationmodel}, $\mathbf{r}_i$ is \textit{She has $16-3=13$ eggs left. So she has $16*2-3=\$13$.}, and $\mathbf{a}_i$ is $\$13$.
Then we calculate the answer distribution for the answer set $\{\mathbf{a}_{i, \cdots, N}\}$ of $\mathbf{s}$. For each distinct value $\mathbf{a} \in \{\mathbf{a}_{i, \cdots, N}\}$, the probability is as follows:
\begin{equation}
\mathbf{p}_{\mathbf{a}} = \frac{{\textstyle \sum_{i=1}^{ | N |}} \mathbf{1}  (\mathbf{a}_i = \mathbf{a})}{ | N | }
\label{equation:answer_probability}
\end{equation}
where $|N|$ denotes the number of answers. For example, in Fig.~\ref{fig:illustrationmodel}, there are three answers as well as three rationales.
We use the answer entropy as the Diversity Entropy (DE) score for the sample $\mathbf{s}$:

\begin{equation}
\mathbf{DE} = \sum_{\mathbf{a}\in \{\mathbf{a}_i\}}^{} -\mathbf{p}_{\mathbf{a}} \log_{}{\mathbf{p}_{\mathbf{a}}}  
\end{equation}
The higher the DE score, the more likely it needs manual correction.
A threshold $\mathbf{\alpha}$ is set for DE as the hyper-parameter.

\subsection{Correction Stage}\label{subsection:manual_correction}
Referring to how humans should involve in the loop, the most straight-forward idea is humans handling the filtered samples while the LLM processes the rest samples. However, humans handling the sample as a whole problem is still labor-consuming, especially for those difficult mathematical problems. Due to this, we claim that humans should pay local attention to simple sub-logics in the rationale. Here, a sub-logic is typically a group of words that can stand alone as a complete thought in a complex rationale. We denote a sentence as a sub-logic. 

To support our claim, there exist some premises.
Firstly, an incorrect rationale could output the correct final answer after correcting the erroneous sub-logic in the rationale. To empirically prove that, we conduct quantitative experiments for twelve datasets and discover that in general up to 50$\%$ of errors of CoT indeed are caused by incorrect intermediate rationales. After correcting these 50$\%$ incorrect rationales, the final answers turn out to be correct. Secondly, correcting sub-logics indeed solves the majority of incorrect rationales. We conduct the analytical experiment across multiple tasks in Sec.~\ref{subsection:error_case_study} and provide the evidence. Thirdly, the questionnaire survey shows that correcting each sub-logic independently is much easier and more user-friendly for humans than checking the entire rationale (Please refer to Appendix ~\ref{appendix:experiments_for_correction_stage} for more details).

Specifically, in the correction stage, we ask humans to check the filtered sample and only correct the rationale with the highest probability. 
During the correction, to simplify, the operations that a human can perform on the sub-logics include ``modifying'', ``adding'', and ``deleting''.
As shown in Tab.~\ref{table:human_corrction}, the first cause displays the modifying operation. 
After the modifying operation, the corrected sub-logic ``\textit{$3 * 100 + 8 * 10 + 3 * 1 = 383$}'' helps the LLM output the correct answer. 


\begingroup
\begin{table}[ht]
    \normalsize
    \small
    \resizebox{\columnwidth}{!}{
    \begin{tabular}{p{\linewidth}}
        \toprule
        \textbf{Correction Operation:} Modifying \\
        \midrule
        \textbf{\textsc{Question:}} 
        Q: I have 3 hundred, 8 tens, and 3 ones. What number am I? A: \\
        \textbf{\textsc{Rationale:}}
        I have 3 hundred, 8 tens, and 3 ones. That means I have \textbf{<<Before Modifying>>:} \textcolor{red}{ $3 * 100 + 8 * 10 + 3 * 1 = 303$ } \textbf{ <<After modifying>>:} \textcolor[HTML]{3CB371}{ $3 * 100 + 8 * 10 + 3 * 1 = 383$}. \\
        \midrule
        \textbf{Correction Operation:} Deleting \\
        \midrule
        \textbf{\textsc{Question:}} 
        Clarence has 5 oranges. He gets 3 more from Joyce. Later, Clarence buys 9 Skittles at the store. How many oranges does Clarence have in all? A: \\
        \textbf{\textsc{Rationale:}} 
        Clarence has 5 oranges. He gets 3 more from Joyce, so now he has $5 + 3 = 8$ oranges. \textbf{<<Delete>>:} \textcolor{red}{ Later he buys 9 Skittles at the store, so he has $8 - 9 = -1$ oranges.} \\
        \midrule
        \textbf{Correction Operation:} Adding \\
        \midrule
        \textbf{\textsc{Question:}} 
        Q: There are 83 trees in a park. 36 of them are willows and the rest are oaks. How many more oaks than willows are there in the park? A: \\
        \textbf{\textsc{Rationale:}} 
        There are 83 trees in the park. 36 of them are willows, and the rest are oaks. This means there are $83 - 36 = 47$ oaks in the park. There are 47 more oaks than willows. \textbf{<<Add>>:} \textcolor[HTML]{3CB371}{ There are 36 willows and 47 oaks in the park now, so there are $47 - 36 = 11$ more oaks than willows.} \\
        \bottomrule
    \end{tabular}
    }
    \caption{
    Examples of manual correction for incorrect sub-logic. The operations that a human can perform on the rationales include modifying, adding, and deleting.
    }
    \label{table:human_corrction}
    \vspace{-5mm}
\end{table}
\endgroup

\section{Cost-utility Analysis Model for Human-in-the-Loop Systems}
\label{sec:framework}

CAMLOP introduces the cost-utility relation that is introduced from the economics theory~\citep{intermediate_microeconomics} into the AI field to quantify these two factors (\ie, cost and utility). For human-in-the-loop systems like \model, we divide the goods into two simple categories: human labor and LLM. Company strategic decision-makers always choose the best bundle of goods they can afford/cost. 
The costs include direct and indirect costs. The direct cost is the money the goods spent while indirect costs mainly include overhead costs from management and rent. Indirect costs also include intangible costs, such as the impact on customers, employees, or delivery times should be considered. Utilities include boosted accuracy, social prestige, and user satisfaction. For simplicity, we only consider money and time for cost while considering accuracy and user satisfaction for utility in our experiments.
\begin{wrapfigure}[13]{r}{0.3\textwidth}
  \centering
    \includegraphics[width=0.3\textwidth]{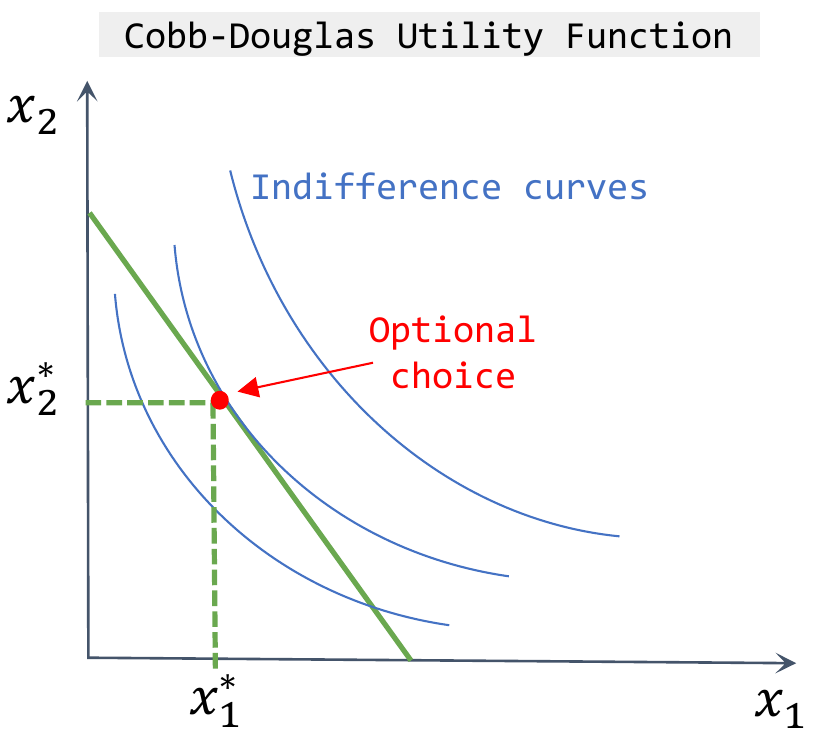}
    
    \caption{Illustration of CAMLOP.}
    \label{fig:cail}
\end{wrapfigure}

We draw Fig.~\ref{fig:cail} where the horizontal axis $x_1$ and vertical axis $x_2$ are the quantity of human labor and LLMs respectively. First, we introduce notations related to the cost. We define $p_1 * x_1$ as the cost spent on human labor and $p_2 * x_2$ as the cost spent on the LLMs.
We indicate the bundle by $(x_1, x_2)$ (a data point in Fig.~\ref{fig:cail}). The corresponding unit price is $p_1$ and $p_2$. The total cost the company decision-maker has to spend is denoted as $y$. Therefore, the budget constraint can be represented as $p_1 x_1 + p_2 x_2 \leq m$. The \textcolor[rgb]{0.28,0.49,0.08}{solid straight line} is the set of data points that cost exactly $y$: $p_1 x_1 + p_2 x_2 = m$.
To note, the cost contains various aspects as mentioned before. In Fig.~\ref{fig:cail}, for simplicity, we express these different aspects as a unified value according to a unified standard.
Then we introduce utilities~\footnote{Most notations are following those from \citep{intermediate_microeconomics}}. 
A utility function $u(x_1, x_2)$ is a way to assign a utility value to the bundle $(x_1, x_2)$. 
As shown in Fig.~\ref{fig:cail}, the set of all data points $(x_1, x_2)$ such that $u(x_1, x_2)$ equals a constant is called a level set (\textcolor[rgb]{0.4,0.5,0.99}{solid curve}). Those data points on higher indifference curves are getting larger utility. We adopted a commonly used utility function--- Cobb-Douglas\footnote{\url{http://www.columbia.edu/~md3405/IM_recap_1_16.pdf}} utility function $u(x_1, x_2) = x_1^c x_2^d$, where $c$ and $d$ are positive numbers that we need to learn~\footnote{Cobb-Douglas indifference curves is what economists referred to as ``well-behaved indifference curves''. Cobb-Douglas utility functions are proved useful to present algebraic examples of the economic field.}. 
Given a model parameterized by $c, d$, and a fixed cost $y$, the model predicts the \textcolor{red}{optimal choice} $(x_1^{*}, x_2^{*})$ with the highest utility, which is desired by the company strategic decision-makers.
Note an important feature of this optimal choice: at this data point the indifference curve is tangent to $p_1 x_1 + p_2 x_2 = y$.

To note, we introduce the modeling of CAMLOP in this section. More details about the inference and learning are shown in Appendix ~\ref{appendix:inference} and Appendix ~\ref{appendix:learning}.

\section{Experiments}
\label{sec:experiments}

\subsection{Setup}

\paragraph{Tasks and datasets.} 
For arithmetic reasoning tasks, we conducted a series of experiments on the Math Word Problem Repository \citep{amini2019mathqa}, including AddSub \citep{hosseini2014learning}, MultiArith \citep{roy2016solving}, SingleEq \citep{koncel2015parsing} and SingleOp \citep{kushman2014learning}. We also included  ASDiv \citep{miao2021diverse}, AQUA-RAT \citep{miao2021diverse}, GSM8K \citep{cobbe2021training}, and ASDiV \citep{patel2021nlp}.
For commonsense reasoning tasks, we used CommonsensQA\citep{talmor2018commonsenseqa} and StrategyQA\citep{geva2021did}. For symbolic reasoning tasks, we used Last Letter Concatenation and Coinflip\citep{wei2022chain}

\paragraph{Baselines.} 
We primarily compare \model with the following baselines. It is noteworthy that all baselines use the same LLM as the decoder. 
For a fair comparison, we report the results of Self-consistency, \model, and \model + Self-consistency with the same 5 rationales sampled from the decoder. The details of the baselines are as follows:
\begin{enumerate}[]
\item \textit{CoT-prompting.} Chain-of-thought prompting with greedy decoding~\citep{wei2022chain}.
\item \textit{Self-consistency.} Chain-of-thought prompting replacing the greedy decoding strategy used in CoT-prompting. Self-consistency generates a set of rationales by sampling from LLM's decoder and determines the optimal answer by taking a majority vote ~\citep{wang2022self}.
\end{enumerate}

\paragraph{Models and scales.} 
We use GPT-3 \citep{ouyang2022training,brown2020language}
\footnote{The \texttt{text-davinci-002} version is InstructGPT. We use the \textit{text-davinci-002} version of GPT-3 to finish all the experiments.} 
with 175-billion parameters as the LLM. 
More details are provided in Appendix ~\ref{appendix:experiment_details}.
For our methods, we provide the following two variants:
\begin{enumerate}
\item \textit{\model.} \model is the result of manual correction for the top $40\%$ CoT predictions ranked out using DE.
A detailed analysis of the threshold of Diversity Entropy is shown in Sec.~\ref{paragraph:ablation-diversity-metrics-ratio}.
\item \textit{\model+Self-consistency.} \model + Self-consistency is the result of combining marginalizing out the sampled rationales with \model. In practice, we use Self-consistency to get answers by majority vote, and then we use \model to manually correct incorrect sub-logics of the first rationale out of decoded rationales with DE calculated based on the decoded rationales.
\end{enumerate}

\paragraph{Sampling scheme.}  To sample diverse rationales, we followed similar settings to those used in ~\citet{wang2022self} for the open-text generation. We use $T = 0.7$ without top-$k$ truncation. For a fair comparison, we use the same prompts as in \citet{wei2022chain}. The threshold of DE is set to be top $40\%$

\subsection{Main Results}\label{sec:main_results}



\paragraph{Arithmetic Reasoning} The results are shown in Tab.~\ref{tab:main-results-arithmetic}. \model generally improves the arithmetic reasoning performance at a large margin ($4.68$ points on average) compared with CoT. \model + Self-consistency further improves the arithmetic reasoning performance ($6.39$ points on average). Especially for SingleEq and SVAMP, compared with CoT, the accuracy increased by 9.05 and 12.10 points, respectively. \model + Self-Consistency performs 

\begin{table*}[ht]
\centering
\small
\setlength\tabcolsep{3pt}
    \centering
        \begin{tabular}{ccccccccc}
    \toprule
     & AddSub & MultiArith & SingleEq & SingleOp & ASDiv & AQuA & SVAMP & GSM8K\\
    \midrule
    CoT-prompting & 82.78 & 93.00 & 85.04 & 94.84 & 73.19 & 40.55 & 68.00 & 56.48 \\
    Self-consistency & 90.63 & 94.17 & 89.17 & 95.73 & 77.72 & 38.19 & 75.70 & 58.85 \\
    \midrule
    \model & 92.15 & \textbf{95.50} & 92.51 & 96.62 & 75.52 & \textbf{44.09} & 74.60 & 61.56 \\
    \model + Self-consistency & \textbf{97.22} & \textbf{95.50} & \textbf{94.09} & \textbf{98.75} & \textbf{79.63} & 41.34 & \textbf{80.10} & \textbf{62.92} \\
    \bottomrule
    \end{tabular}
    \caption{Arithmetic reasoning accuracy by \model and \model + Self-consistency compared to Chain-of-Thought prompting and Self-consistency. For each task, we report the median scores among 5 runs. }
    \label{tab:main-results-arithmetic}
    \vspace{-5mm}
\end{table*}

\paragraph{Commonsense and Symbolic Reasoning} Tab.~\ref{tab:main-results-commonsense} shows the results on commonsense and symbolic reasoning tasks. Similarly, \model improves the performance and \model + Self-consistency further boosts it. 
For symbolic reasoning, we adopt the out-of-distribution (OOD) setting where the input prompt contains samples of 4-letters and 4-flips~ \citep{wang2022self} because this setting is more challenging. We do not adopt the in-distribution setting because GPT-3 can already achieve 100\% accuracy with the in-distribution setting as shown in \citet{wei2022chain}. Even in difficult OOD setting, the gain of \model+Self-consistency is significant compared to CoT-prompting and Self-consistency.

\begin{table}[htbp]
\centering
\small
\setlength\tabcolsep{3pt}
    \centering
    \begin{tabular}{ccccc}
    \toprule
    \multirow{2}{*}{Model} & \multicolumn{2}{c}{\textit{Commonsense}} & \multicolumn{2}{c}{\textit{Symbolic}} \\
    \cmidrule(r){2-3}
    \cmidrule(r){4-5}
     & CSQA & StraQA & Letter & Coinflip\\
    \midrule
    CoT-prompting & 72.32 & 60.13 & 49.20 & 81.40\\
    Self-consistency & 76.09 & 61.40 & 54.40 & \textbf{93.20}\\
    \midrule
    \model & 73.71 & 60.88 & 75.40 & 81.40\\
    \model + Self-consistency & \textbf{77.07} & \textbf{62.23} & \textbf{78.40} & \textbf{93.20}\\
    \bottomrule
    \end{tabular}
    \caption{Commonsense and symbolic reasoning accuracy. For each task, we report the median scores among 5 runs. }
    \label{tab:main-results-commonsense}
    \vspace{-5mm}
\end{table}


\begin{figure*}[ht]
\centering
\includegraphics[width=1.0\linewidth]{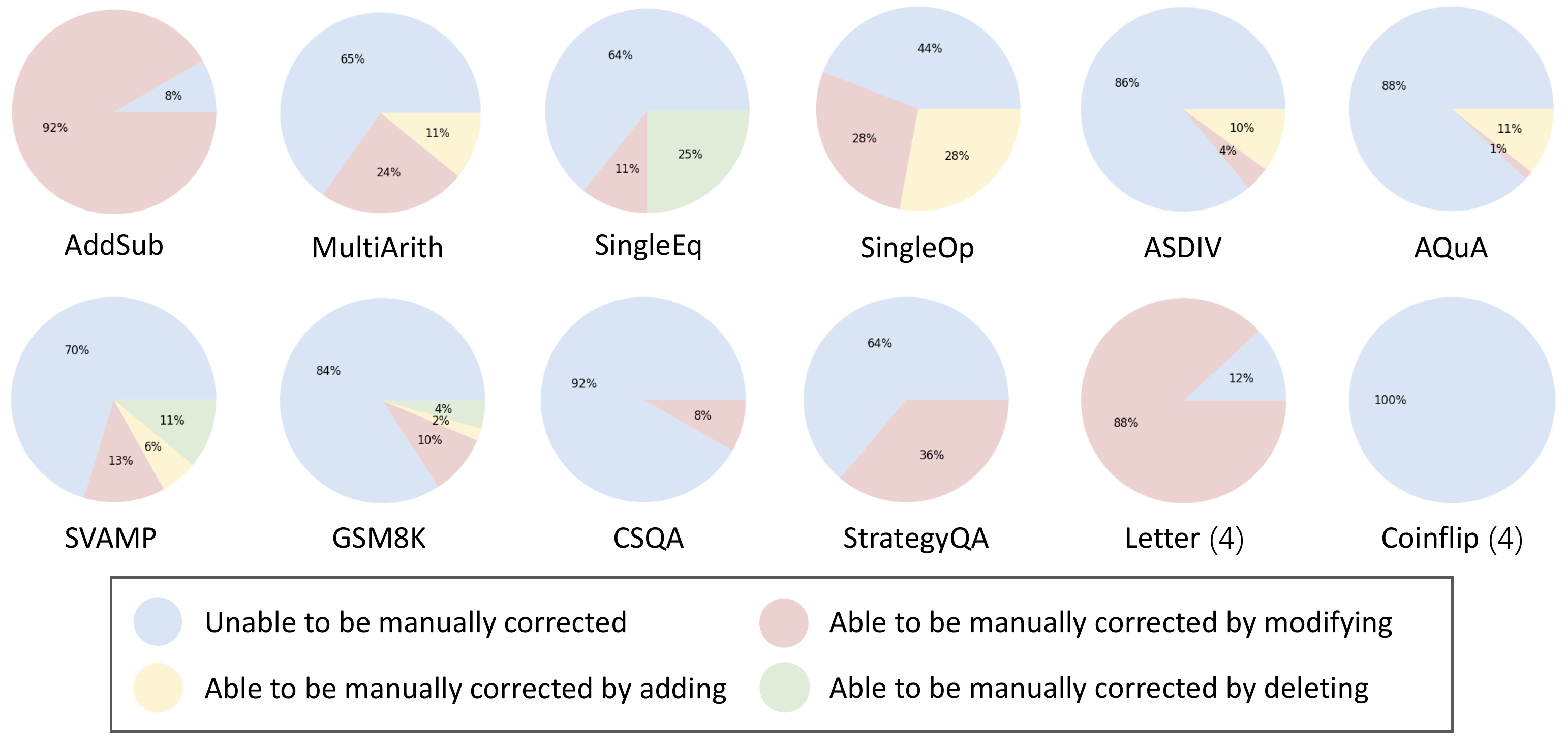}
\caption{
Illustration of error analysis of Chain of Thought Prompting across twelve tasks. Each error type is represented by a color. The share in color indicates the share of the error type.
}
\label{fig:error_case_study}
\vspace{-5mm}
\end{figure*}

\subsection{Analysis of Whether Correcting Sub-logics Solves the Majority of Incorrect Rationales}
\label{subsection:error_case_study}

We conduct experiments on twelve datasets to check whether correcting sub-logics solves the majority of incorrect rationales. Each task is represented by a pie chart. For each task, we conduct the error analysis for CoT prompting and analyze the error types of rationales. We divided the error types into four categories: errors that are able to be corrected by the \textcolor[HTML]{FA8072}{``modifying''} operation, the \textcolor[HTML]{FFBE1E}{``adding''} operation, the \textcolor[HTML]{3CB371}{``deleting''} operation, and the rest of the errors that are \textcolor[HTML]{87C3FA}{unable to be manually corrected}. The percentage of each type across datasets is shown in Fig.~\ref{fig:error_case_study}. More details are shown in Appendix ~\ref{appendix:error_case_study}.

The first three categories constituent the majority of incorrect rationales and can be solved by correcting independent sub-logics instead of the whole rationale. 
More specifically, CoT often makes mistakes when calculating polynomial calculations with decimal points, which account for a large part of manual correction and can be corrected by the \textcolor[HTML]{FA8072}{``modifying''} operation. For the \textcolor[HTML]{FFBE1E}{``adding''} operation, it functions when CoT often fails to convert the units, for example, from grams to kilograms.
CoT often outputs redundant logic, leading to incorrect answers, which could be fixed by the \textcolor[HTML]{3CB371}{``deleting''} operation.
Except for the error mentioned above, errors that are \textcolor[HTML]{87C3FA}{unable to be manually corrected} include misinterpretation of the question, incorrect formula, whole incorrect composition of sub-logics and so on.

Additionally, we find that the advantage of Self-consistency often comes from fixing the errors that are \textcolor[HTML]{87C3FA}{unable to be manually corrected}. Sampling a large set of rationales and taking a majority vote helps the fix of misinterpretation of the question while making little help in fixing calculation error.
On the contrary, \model is beneficial for other three categories of errors including \textcolor[HTML]{FA8072}{``modifying''}, \textcolor[HTML]{FFBE1E}{``adding''} and \textcolor[HTML]{3CB371}{``deleting''}.
The difference between Self-consistency and \model illustrates why \model + Self-consistency achieves great performance as shown in Tab. ~\ref{tab:main-results-arithmetic}. Obviously, \model and Self-consistency play different roles and be mutually complementary.

\subsection{Additional Study}
\label{section:additional_study}

\paragraph{Validation of Diversity Entropy}
\label{paragraph:ablation-diversity-metrics-ratio}

\begin{figure}
\centering
\includegraphics[width=1.0\linewidth]{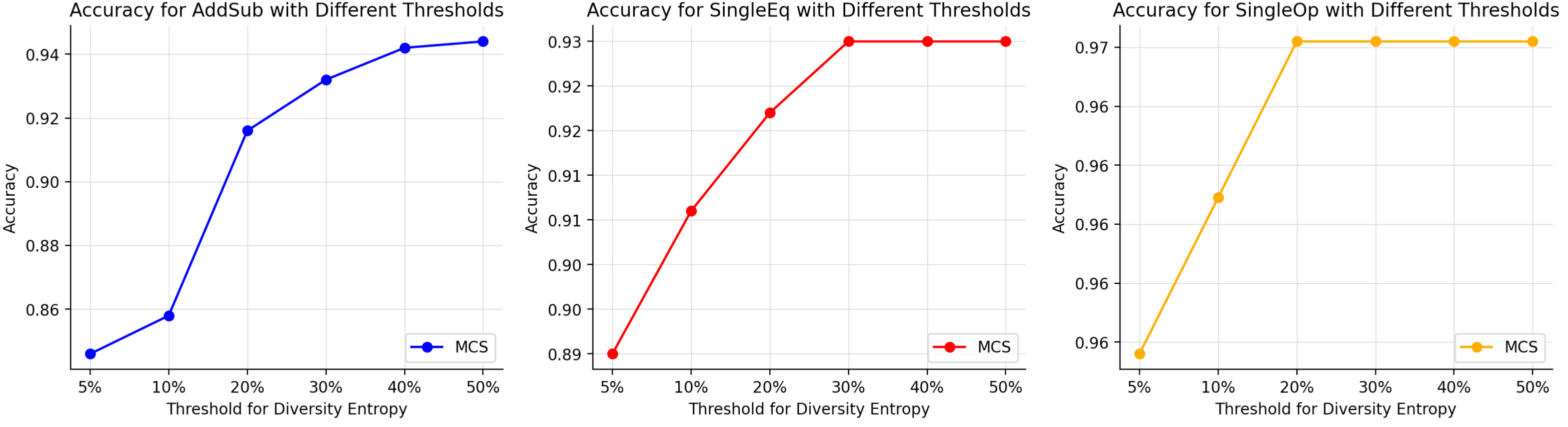}
\caption{
Results of different thresholds of DE. It shows the results of \model with 5\%, 10\%, 20\%, 30\%, 40\% and 50\% DE for \textcolor{blue}{AddSub} (Left), \textcolor{red}{SingleEq} (Medium) and \textcolor[HTML]{FF9B1E}{SingleOp} (Right). Results show that DE-based filtering is an efficient method to rank the possibility to be incorrect for the output of CoT predictions, and samples with incorrect output will be ranked higher than those without.
}
\label{fig:threshold}
\vspace{-4mm}
\end{figure}

\begin{figure}
\centering
\includegraphics[width=1.0\linewidth]{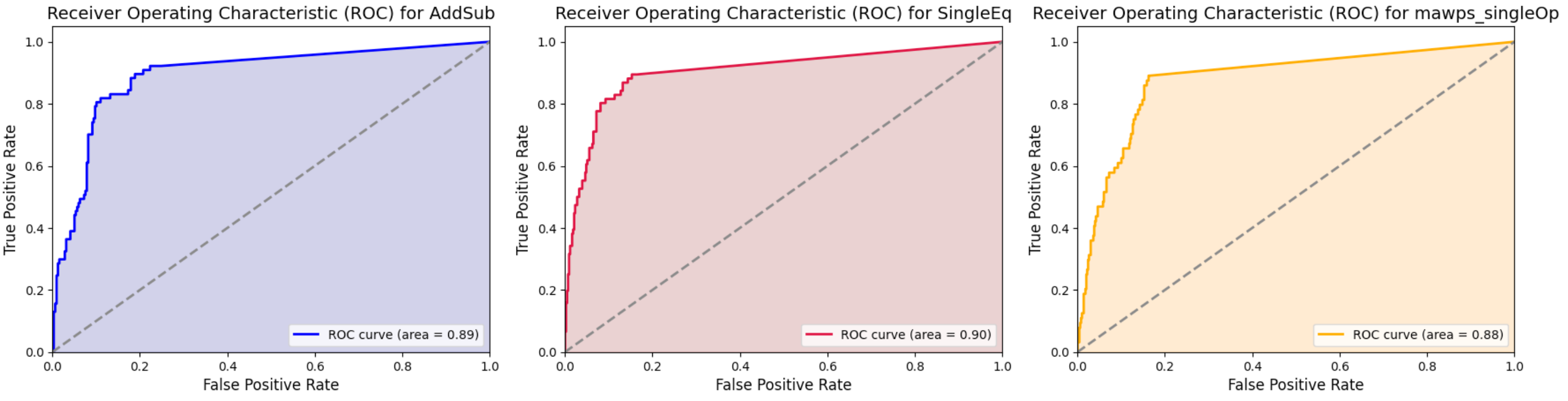}
\caption{
ROC Curves for DE to filter out the incorrect CoT outputs. It shows the ROC Curve for \textcolor{blue}{AddSub} (Left), \textcolor{red}{Singleeq} (Medium) and \textcolor[HTML]{FF9B1E}{SingleOp} (Right). The results indicate that DE is a reliable metrics that can determine the samples most likely to be incorrectly predicted for humans to involve.
}
\label{fig:ROCAUC}
\vspace{-6mm}
\end{figure} 

To validate the effectiveness of Diversity Entropy in determining whether the manual correction is necessary for each sample, we draw a ROC Curve in Fig.~\ref{fig:ROCAUC} to demonstrate its ability to rank the likelihood of incorrect outputs.
The selection of the threshold involves a trade-off between performance and human labor.
Fig.~\ref{fig:threshold} shows that the performance stabilizes after reaching the threshold of top $20\%$ to top $40\%$ for most datasets. Therefore, we set the threshold to be top $40\%$ across all our experiments.
As the manual correction is labor-consuming and time-consuming, Diversity Entropy can help save time and labor by allowing humans to focus on checking only a small percentage.

\paragraph{Analysis of Aggregation Strategies}
\label{paragraph:diversity_metrics_over_diverse_reasoning_paths}

\begin{table}[ht]

\small
\setlength\tabcolsep{4pt}
    \centering
    \begin{tabular}{l cccccc}
    \toprule
        Calculation Strategy  & ASDiv & AQuA & SVAMP & GSM8K\\
       \midrule
        Unnormalized Weighted Average & 73.71 & 44.09 & 74.50 & 61.41 \\
       Normalized Weighted Average & 73.71 & 40.94 & \textbf{74.60} & \textbf{61.56} \\
       \midrule
       Unnormalized Weighted Sum & 73.80 & 42.52 & 74.50 & 60.20 \\
       Normalized Weighted Sum & 73.37 & \textbf{44.88} & 71.30 & 59.21 \\
       \midrule
       Unnormalized Unweighted Sum (Majority Vote) & \textbf{75.52} & 44.09 & \textbf{74.60} & \textbf{61.56} \\
       \bottomrule
    \end{tabular}
    \caption{Accuracy comparison of different strategies of computing answer probability. The threshold of Diversity Metrics is set to be top $40\%$. 
    }
    \label{table:aggregation}
    \vspace{-5mm}
\end{table}

The majority vote method of calculating the answer probability over all sampled rationales can be regarded as taking an unnormalized unweighted sum. As described in \citet{wang2022self}, other methods of computing answer probability of $\mathbf{a}$ include the unnormalized weighted average, normalized weighted average, unnormalized weighted sum, 
\begin{wrapfigure}[14]{r}{0.35\textwidth}
\centering
\includegraphics[width=1.0\linewidth]{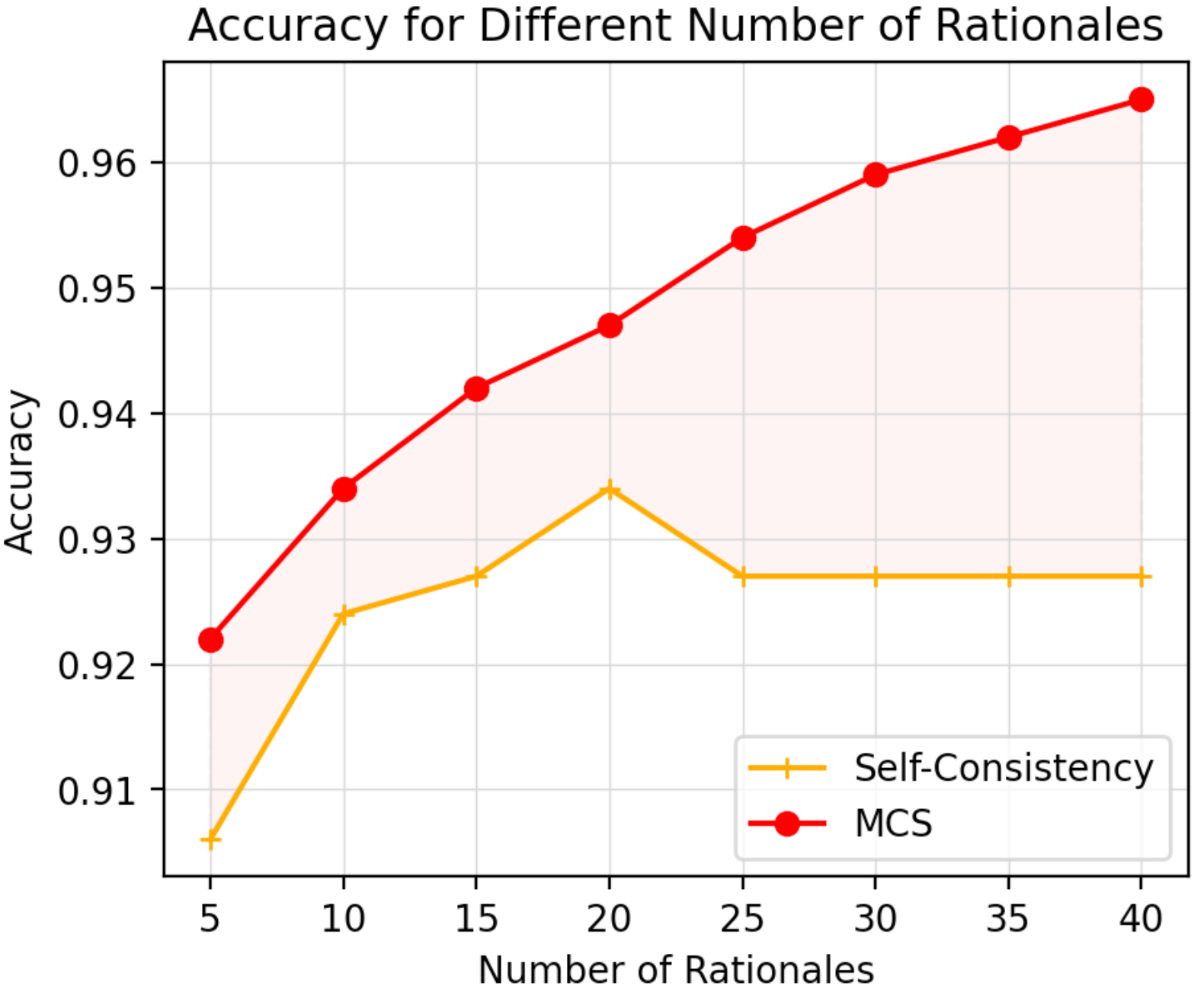}
\caption{
Experiments of different numbers of rationales.}
\label{fig:numbers}
\end{wrapfigure}
and normalized weighted sum. More details about the above calculation are provided in Appendix ~\ref{https://www.overleaf.com/project/6454727bfead9dd3cc65bd0a}. 
Tab.~\ref{table:aggregation} shows that unnormalized unweighted sum generally outperforms others. We use this setting in all experiments following~\citet{wang2022self}.


\paragraph{Analysis of the Number of Sampled Rationales}

We test the accuracy with respect to varying the number of rationales (\ie, $5$, $10$, $15$, $20$, $25$, $30$, $35$, $40$) in Fig.~\ref{fig:numbers}.  The results are arithmetic reasoning accuracy on SingleEq. For a fair comparison, both \model and Self-consistency use the same prompts as in \citet{wei2022chain}. Both \textcolor{red}{\model} and \textcolor[HTML]{EAA629}{Self-consistency} use the same $5$ rationales sampled from the decoder. In our experiments, the threshold of Diversity Metrics is set to be top $40\%$. The results show that \model generally outperforms self-consistency and benefits from the increasing number of sampled rationales.

\subsection{Balancing Cost and Utility}\label{sec:analysis_cost_utility}

\begin{table}[ht]
\centering
\small
\setlength\tabcolsep{3pt}
    \centering
        \begin{tabular}{lcccc}
    \toprule
    Plans & Time & Money & Acc. & Utility(User Satis.) \\
    \midrule
    Human & $60$s & $\$0.125$ & 93.20 & 86.40 \\
    \midrule
    CoT Prompting & $0.8$s & $\$0.080$ & 85.04 & 81.60 \\
    Self-Consistency ($\mathbf{N}_{self} = 10$) & $8$s & $\$0.800$ & 92.49 & 85.80 \\
    \midrule
    \model ($\mathbf{N}_{\model} = 5$, $\mathbf{\alpha} = 20\%$) & $10.8$s & $\$0.4925$ & 91.00 & 84.20 \\
    \model + Self-consistency ($\mathbf{N}_{\model} = 5$, $\mathbf{\alpha} = 20\%$) & $10.8$s & $\$0.4925$ & 93.50 & 88.80 \\
    \midrule
    \model ($\mathbf{N}_{\model} = 5$, $\mathbf{\alpha} = 40\%$) & $16.8$s & $\$0.505$ & 92.51 & 85.60 \\
    \model + Self-consistency ($\mathbf{N}_{\model} = 5$, $\mathbf{\alpha} = 40\%$) & $16.8$s & $\$0.505$ & 94.09 & 90.80 \\ 
    \bottomrule
    \end{tabular}
    \caption{Analysis of cost and utility for SingleEq.
    \model + Self-consistency generally outperforms other methods with higher utility and acceptable cost. $\mathbf{N}_{\cdot}$: \# sampled rationale. $\mathbf{\alpha}$: DE threshold. Acc.: Accuracy. User Satis.: User Satisfaction. More details are shown in Appendix ~\ref{appendix:details_of_balancing_cost_and_utility}.}
    \label{tab:utility-cost}
    \vspace{-5mm}
\end{table}

In this section, we conduct experiments on the SingleEq dataset to quantitatively calculate cost and utility for \cail.
For the cost, we consider money and time.
We set the price of the LLM as $\mathbf{p}_{llm}$ and the time cost as $\mathbf{t}_{llm}$. Since we use GPT-3, the price $\mathbf{p}_{llm}$ for a single math problem (decoding once) is $\$0.08$ on average, and the time cost $\mathbf{t}_{llm}$ is $0.8$ second based on empirical results
\footnote{The pricing of \texttt{text-davinci-002} is $\$0.02$ per $1000$ tokens, which can be found at  \url{https://openai.com/pricing}. We set $\mathbf{p}_{llm}$ to be $\$0.08$ because an input sample for few-shot CoT contains about $4000$ tokens on average when decoding only once. Note that we only calculated the time for the main part (\ie, the decoding) and ignored other parts that were fast enough to be ignored compared to the API calls.}.
The price of solving a single math problem with only human labor is $\mathbf{p}_{human}$ and the time cost is $\mathbf{t}_{human}$.  We set $\mathbf{p}_{human}$ to be $\$0.125$ and $\mathbf{t}_{human}$ to be $60$ seconds based on our empirical results.
\footnote{Minimum hourly wage in the United States is $\$7.5$, which can be found at \url{https://www.worker.gov/pay-for-hours-worked/}. Solving a problem requires $60$ seconds on average. Therefore, the price and time cost required to complete a problem are $\$0.125$ and $60$ seconds, respectively.}
The price of human labor for \model to correct a single math problem $\mathbf{p}_{\model}$ is $\$0.0625$ and the time cost $\mathbf{t}_{\model}$ is $30$ seconds based on empirical results. Note the time required to inspect and correct is less than the time needed to fully solve the entire problem, therefore $\mathbf{t}_{\model} < \mathbf{t}_{human}$.

For the utility, we consider user satisfaction as the comprehensive score. We ask five users to write down their satisfaction levels and calculate the average~\footnote{See Appendix for more details about user satisfaction. The impact of accuracy on user satisfaction is much larger than the time cost, we speculate that most users care more about the accuracy of solving the problem than the time cost, as SingleEq is a math-solving dataset.}. 
We also perform regression analysis on user satisfaction based on LLM and Human and ultimately learn the utility function $\mathbf{u}(\mathbf{x}_{llm}, \mathbf{x}_{human}) = \mathbf{x}_{llm}^{2.05}*\mathbf{x}_{human}^{1.94}$. For more details, please refer to Appendix ~\ref{appendix:details_of_balancing_cost_and_utility}.

We experiment on five candidate plans based on models from Sec.~\ref{sec:main_results} and Sec. \ref{section:additional_study} (Fig.~\ref{fig:threshold} and Fig.~\ref{fig:numbers}):
\begin{enumerate}[]
\item \textit{Human}: A plan that requires only human labor, which costs $\mathbf{p}_{human}$ and $\mathbf{t}_{human}$ seconds.
\item \textit{CoT-prompting}: A naive CoT plan that only requires GPT-3 for decoding only once, which costs $\mathbf{p}_{llm}$ and $\mathbf{t}_{llm}$ seconds.
\item \textit{Self-consistency}: A Self-consistency plan that requires only LLMs to sample from the decoder $\mathbf{N}_{self}$ times, which will cost $ \mathbf{N}_{self} * \mathbf{p}_{llm}$ and $\mathbf{N}_{self} * \mathbf{t}_{llm} $ seconds.
\item \textit{\model}: \model samples from LLM decoder $\mathbf{N}_{\model}$ times and uses top $\mathbf{\alpha}$ as threshold, requiring $ (\mathbf{N}_{\model}+1) * \mathbf{p}_{llm} + \mathbf{\alpha} * \mathbf{p}_{\model}$ and $ (\mathbf{N}_{\model}+1) * \mathbf{t}_{llm} + \mathbf{\alpha} * \mathbf{t}_{\model}$ seconds.
\item \textit{\model + Self-consistency}: A \model + Self-consistency plan that requires to sample from the decoder $\mathbf{N}_{\model}$ times, which costs the same as the \model plan.
\end{enumerate}

The results are shown in Tab.~\ref{tab:utility-cost}. The result shows that \model+Self-consistency generally outperforms other methods with higher utility (\ie, better user satisfaction) as well as an acceptable cost.

\section{Related Work}
\label{section:related_work}

The human-in-the-Loop system, aiming to achieve what neither humans nor machines can accomplish independently, is defined as a model requiring human interaction~\citep{karwowski2006international}. When the machine cannot solve the problem, or when cost or security considerations require humans to participate, manual intervention is necessary~\citep{bien2018deep,wu2022survey, zanzotto2019human, mosqueira2023human}. Previous human-in-the-loop systems focus either on adding appropriate tags to data or providing feedback on cases with a certain confidence interval to the machines and thus retrain the model afterward with the labeled data or rewarded cases~\citep{wu2022survey,zanzotto2019human}. 

Recently, LLM-based AI (Artificial Intelligence) systems are developing very quickly, and this trend is expected to expand to the majority of the workforce in the near future~\citep{ouyang2022training, zhang2022opt, sanh2021multitask}. However, these systems do not always provide satisfactory answers without human intervention. Additionally, in domains such as criminal fact identification and charge predictions, inference should be reasonable and controlled by humans~\citep{custers2022ai} while LLMs are not qualified.
Therefore, it is essential to develop a human-in-the-loop prompting-based system that is designed with the ability to collaborate with humans. Until recently, few researchers have systematically and quantitatively explored human-in-the-loop prompting-based systems.

Different from ChatGPT's RLHF (\ie, Reinforcement Learning from Human Feedback)~\footnote{\url{https://openai.com/blog/chatgpt}.}, we take the first step to use human feedback in an online way without access to parameters. Even though it's a preliminary step, this online method could benefit from further refinement and combination with RLHF in future research.

\section{Conclusion}
We propose the MCS to explore how manual correction of rationales can improve LLM's reasoning ability. Then, we propose CAMLOP to quantitatively and systematically analyze and balance the cost and the corresponding utility. Experiments demonstrate that our \model significantly outperforms strong baselines including the CoT prompting approach and Self-consistency approach and obtains the optimal balance between cost and utility. 

\bibliographystyle{plainnat}
\bibliography{main.bib}

\begin{thebibliography}{38}
\providecommand{\natexlab}[1]{#1}
\providecommand{\url}[1]{\texttt{#1}}
\expandafter\ifx\csname urlstyle\endcsname\relax
  \providecommand{\doi}[1]{doi: #1}\else
  \providecommand{\doi}{doi: \begingroup \urlstyle{rm}\Url}\fi

\bibitem[Ackley et~al.(1985)Ackley, Hinton, and Sejnowski]{ackley1985learning}
David~H Ackley, Geoffrey~E Hinton, and Terrence~J Sejnowski.
\newblock A learning algorithm for boltzmann machines.
\newblock \emph{Cognitive science}, 9\penalty0 (1):\penalty0 147--169, 1985.

\bibitem[Agarwal et~al.(2020)Agarwal, Arora, Anand, and
  Arora]{agarwal2020contextual}
Sharat Agarwal, Himanshu Arora, Saket Anand, and Chetan Arora.
\newblock Contextual diversity for active learning.
\newblock In \emph{European Conference on Computer Vision}, pages 137--153.
  Springer, 2020.

\bibitem[Amini et~al.(2019)Amini, Gabriel, Lin, Koncel-Kedziorski, Choi, and
  Hajishirzi]{amini2019mathqa}
Aida Amini, Saadia Gabriel, Peter Lin, Rik Koncel-Kedziorski, Yejin Choi, and
  Hannaneh Hajishirzi.
\newblock Mathqa: Towards interpretable math word problem solving with
  operation-based formalisms.
\newblock \emph{arXiv preprint arXiv:1905.13319}, 2019.

\bibitem[{BENJ EDWARDS}(2023)]{IBM}
{BENJ EDWARDS}.
\newblock Ibm plans to replace 7,800 jobs with ai over time, pauses hiring
  certain positions, {IBM CEO Arvind Krishna says he could see 30\% of
  back-office functions replaced by AI over 5 years.}, 2023.
\newblock
  \url{https://arstechnica.com/information-technology/2023/05/ibm-pauses-hiring-around-7800-roles-that-could-be-replaced-by-ai/}.

\bibitem[Bien et~al.(2018)Bien, Rajpurkar, Ball, Irvin, Park, Jones, Bereket,
  Patel, Yeom, Shpanskaya, et~al.]{bien2018deep}
Nicholas Bien, Pranav Rajpurkar, Robyn~L Ball, Jeremy Irvin, Allison Park, Erik
  Jones, Michael Bereket, Bhavik~N Patel, Kristen~W Yeom, Katie Shpanskaya,
  et~al.
\newblock Deep-learning-assisted diagnosis for knee magnetic resonance imaging:
  development and retrospective validation of mrnet.
\newblock \emph{PLoS medicine}, 15\penalty0 (11):\penalty0 e1002699, 2018.

\bibitem[Brinker(2003)]{brinker2003incorporating}
Klaus Brinker.
\newblock Incorporating diversity in active learning with support vector
  machines.
\newblock In \emph{Proceedings of the 20th international conference on machine
  learning (ICML-03)}, pages 59--66, 2003.

\bibitem[Brown et~al.(2020{\natexlab{a}})Brown, Mann, Ryder, Subbiah, Kaplan,
  Dhariwal, Neelakantan, Shyam, Sastry, Askell, Agarwal, Herbert-Voss, Krueger,
  Henighan, Child, Ramesh, Ziegler, Wu, Winter, Hesse, Chen, Sigler, Litwin,
  Gray, Chess, Clark, Berner, McCandlish, Radford, Sutskever, and
  Amodei]{NEURIPS20201457c0d6}
Tom Brown, Benjamin Mann, Nick Ryder, Melanie Subbiah, Jared~D Kaplan, Prafulla
  Dhariwal, Arvind Neelakantan, Pranav Shyam, Girish Sastry, Amanda Askell,
  Sandhini Agarwal, Ariel Herbert-Voss, Gretchen Krueger, Tom Henighan, Rewon
  Child, Aditya Ramesh, Daniel Ziegler, Jeffrey Wu, Clemens Winter, Chris
  Hesse, Mark Chen, Eric Sigler, Mateusz Litwin, Scott Gray, Benjamin Chess,
  Jack Clark, Christopher Berner, Sam McCandlish, Alec Radford, Ilya Sutskever,
  and Dario Amodei.
\newblock Language models are few-shot learners.
\newblock In H.~Larochelle, M.~Ranzato, R.~Hadsell, M.F. Balcan, and H.~Lin,
  editors, \emph{Advances in Neural Information Processing Systems}, volume~33,
  pages 1877--1901. Curran Associates, Inc., 2020{\natexlab{a}}.
\newblock URL
  \url{https://proceedings.neurips.cc/paper/2020/file/1457c0d6bfcb4967418bfb8ac142f64a-Paper.pdf}.

\bibitem[Brown et~al.(2020{\natexlab{b}})Brown, Mann, Ryder, Subbiah, Kaplan,
  Dhariwal, Neelakantan, Shyam, Sastry, Askell, et~al.]{brown2020language}
Tom Brown, Benjamin Mann, Nick Ryder, Melanie Subbiah, Jared~D Kaplan, Prafulla
  Dhariwal, Arvind Neelakantan, Pranav Shyam, Girish Sastry, Amanda Askell,
  et~al.
\newblock Language models are few-shot learners.
\newblock \emph{Advances in neural information processing systems},
  33:\penalty0 1877--1901, 2020{\natexlab{b}}.

\bibitem[Cobbe et~al.(2021)Cobbe, Kosaraju, Bavarian, Hilton, Nakano, Hesse,
  and Schulman]{cobbe2021training}
Karl Cobbe, Vineet Kosaraju, Mohammad Bavarian, Jacob Hilton, Reiichiro Nakano,
  Christopher Hesse, and John Schulman.
\newblock Training verifiers to solve math word problems.
\newblock \emph{arXiv preprint arXiv:2110.14168}, 2021.

\bibitem[Custers(2022)]{custers2022ai}
Bart Custers.
\newblock Ai in criminal law: An overview of ai applications in substantive and
  procedural criminal law.
\newblock \emph{Law and Artificial Intelligence}, pages 205--223, 2022.

\bibitem[Fan et~al.(2018)Fan, Lewis, and Dauphin]{fan2018hierarchical}
Angela Fan, Mike Lewis, and Yann Dauphin.
\newblock Hierarchical neural story generation.
\newblock \emph{arXiv preprint arXiv:1805.04833}, 2018.

\bibitem[Ficler and Goldberg(2017)]{ficler2017controlling}
Jessica Ficler and Yoav Goldberg.
\newblock Controlling linguistic style aspects in neural language generation.
\newblock \emph{arXiv preprint arXiv:1707.02633}, 2017.

\bibitem[Geva et~al.(2021)Geva, Khashabi, Segal, Khot, Roth, and
  Berant]{geva2021did}
Mor Geva, Daniel Khashabi, Elad Segal, Tushar Khot, Dan Roth, and Jonathan
  Berant.
\newblock Did aristotle use a laptop? a question answering benchmark with
  implicit reasoning strategies.
\newblock \emph{Transactions of the Association for Computational Linguistics},
  9:\penalty0 346--361, 2021.

\bibitem[{Google Research}(2023)]{Minerva}
{Google Research}.
\newblock Minerva: Solving quantitative reasoning problems with language
  models, 2023.

\bibitem[Holtzman et~al.(2018)Holtzman, Buys, Forbes, Bosselut, Golub, and
  Choi]{holtzman2018learning}
Ari Holtzman, Jan Buys, Maxwell Forbes, Antoine Bosselut, David Golub, and
  Yejin Choi.
\newblock Learning to write with cooperative discriminators.
\newblock \emph{arXiv preprint arXiv:1805.06087}, 2018.

\bibitem[Holtzman et~al.(2019)Holtzman, Buys, Du, Forbes, and
  Choi]{holtzman2019curious}
Ari Holtzman, Jan Buys, Li~Du, Maxwell Forbes, and Yejin Choi.
\newblock The curious case of neural text degeneration.
\newblock \emph{arXiv preprint arXiv:1904.09751}, 2019.

\bibitem[Hosseini et~al.(2014)Hosseini, Hajishirzi, Etzioni, and
  Kushman]{hosseini2014learning}
Mohammad~Javad Hosseini, Hannaneh Hajishirzi, Oren Etzioni, and Nate Kushman.
\newblock Learning to solve arithmetic word problems with verb categorization.
\newblock In \emph{EMNLP}, pages 523--533. Citeseer, 2014.

\bibitem[Karwowski(2006)]{karwowski2006international}
Waldemar Karwowski.
\newblock \emph{International Encyclopedia of Ergonomics and Human Factors, -3
  Volume Set}.
\newblock Crc Press, 2006.

\bibitem[Kojima et~al.(2022)Kojima, Gu, Reid, Matsuo, and
  Iwasawa]{kojima2022large}
Takeshi Kojima, Shixiang~Shane Gu, Machel Reid, Yutaka Matsuo, and Yusuke
  Iwasawa.
\newblock Large language models are zero-shot reasoners.
\newblock \emph{arXiv preprint arXiv:2205.11916}, 2022.

\bibitem[Koncel-Kedziorski et~al.(2015)Koncel-Kedziorski, Hajishirzi,
  Sabharwal, Etzioni, and Ang]{koncel2015parsing}
Rik Koncel-Kedziorski, Hannaneh Hajishirzi, Ashish Sabharwal, Oren Etzioni, and
  Siena~Dumas Ang.
\newblock Parsing algebraic word problems into equations.
\newblock \emph{Transactions of the Association for Computational Linguistics},
  3:\penalty0 585--597, 2015.

\bibitem[Kushman et~al.(2014)Kushman, Artzi, Zettlemoyer, and
  Barzilay]{kushman2014learning}
Nate Kushman, Yoav Artzi, Luke Zettlemoyer, and Regina Barzilay.
\newblock Learning to automatically solve algebra word problems.
\newblock In \emph{Proceedings of the 52nd Annual Meeting of the Association
  for Computational Linguistics (Volume 1: Long Papers)}, pages 271--281, 2014.

\bibitem[Miao et~al.(2021)Miao, Liang, and Su]{miao2021diverse}
Shen-Yun Miao, Chao-Chun Liang, and Keh-Yih Su.
\newblock A diverse corpus for evaluating and developing english math word
  problem solvers.
\newblock \emph{arXiv preprint arXiv:2106.15772}, 2021.

\bibitem[Mosqueira-Rey et~al.(2023)Mosqueira-Rey, Hern{\'a}ndez-Pereira,
  Alonso-R{\'\i}os, Bobes-Bascar{\'a}n, and
  Fern{\'a}ndez-Leal]{mosqueira2023human}
Eduardo Mosqueira-Rey, Elena Hern{\'a}ndez-Pereira, David Alonso-R{\'\i}os,
  Jos{\'e} Bobes-Bascar{\'a}n, and {\'A}ngel Fern{\'a}ndez-Leal.
\newblock Human-in-the-loop machine learning: A state of the art.
\newblock \emph{Artificial Intelligence Review}, 56\penalty0 (4):\penalty0
  3005--3054, 2023.

\bibitem[Ouyang et~al.(2022)Ouyang, Wu, Jiang, Almeida, Wainwright, Mishkin,
  Zhang, Agarwal, Slama, Ray, et~al.]{ouyang2022training}
Long Ouyang, Jeff Wu, Xu~Jiang, Diogo Almeida, Carroll~L Wainwright, Pamela
  Mishkin, Chong Zhang, Sandhini Agarwal, Katarina Slama, Alex Ray, et~al.
\newblock Training language models to follow instructions with human feedback.
\newblock \emph{arXiv preprint arXiv:2203.02155}, 2022.

\bibitem[Patel et~al.(2021)Patel, Bhattamishra, and Goyal]{patel2021nlp}
Arkil Patel, Satwik Bhattamishra, and Navin Goyal.
\newblock Are nlp models really able to solve simple math word problems?
\newblock \emph{arXiv preprint arXiv:2103.07191}, 2021.

\bibitem[Radford et~al.(2019)Radford, Wu, Child, Luan, Amodei, Sutskever,
  et~al.]{radford2019language}
Alec Radford, Jeffrey Wu, Rewon Child, David Luan, Dario Amodei, Ilya
  Sutskever, et~al.
\newblock Language models are unsupervised multitask learners.
\newblock \emph{OpenAI blog}, 1\penalty0 (8):\penalty0 9, 2019.

\bibitem[Roy and Roth(2016)]{roy2016solving}
Subhro Roy and Dan Roth.
\newblock Solving general arithmetic word problems.
\newblock \emph{arXiv preprint arXiv:1608.01413}, 2016.

\bibitem[Sanh et~al.(2021)Sanh, Webson, Raffel, Bach, Sutawika, Alyafeai,
  Chaffin, Stiegler, Scao, Raja, et~al.]{sanh2021multitask}
Victor Sanh, Albert Webson, Colin Raffel, Stephen~H Bach, Lintang Sutawika,
  Zaid Alyafeai, Antoine Chaffin, Arnaud Stiegler, Teven~Le Scao, Arun Raja,
  et~al.
\newblock Multitask prompted training enables zero-shot task generalization.
\newblock \emph{arXiv preprint arXiv:2110.08207}, 2021.

\bibitem[Shao et~al.(2023)Shao, Cai, Liao, Zheng, Yang,
  et~al.]{shao2023compositional}
Nan Shao, Zefan Cai, Chonghua Liao, Yanan Zheng, Zhilin Yang, et~al.
\newblock Compositional task representations for large language models.
\newblock In \emph{The Eleventh International Conference on Learning
  Representations}, 2023.

\bibitem[Srivastava et~al.(2022)Srivastava, Rastogi, Rao, Shoeb, Abid, Fisch,
  Brown, Santoro, Gupta, Garriga-Alonso, et~al.]{srivastava2022beyond}
Aarohi Srivastava, Abhinav Rastogi, Abhishek Rao, Abu Awal~Md Shoeb, Abubakar
  Abid, Adam Fisch, Adam~R Brown, Adam Santoro, Aditya Gupta, Adri{\`a}
  Garriga-Alonso, et~al.
\newblock Beyond the imitation game: Quantifying and extrapolating the
  capabilities of language models.
\newblock \emph{arXiv preprint arXiv:2206.04615}, 2022.

\bibitem[Talmor et~al.(2018)Talmor, Herzig, Lourie, and
  Berant]{talmor2018commonsenseqa}
Alon Talmor, Jonathan Herzig, Nicholas Lourie, and Jonathan Berant.
\newblock Commonsenseqa: A question answering challenge targeting commonsense
  knowledge.
\newblock \emph{arXiv preprint arXiv:1811.00937}, 2018.

\bibitem[Varian(2014)]{intermediate_microeconomics}
Hal~R. Varian.
\newblock \emph{Intermediate microeconomics: a modern approach}.
\newblock New York :W.W. Norton Company, 2014.

\bibitem[Wang et~al.(2022)Wang, Wei, Schuurmans, Le, Chi, and
  Zhou]{wang2022self}
Xuezhi Wang, Jason Wei, Dale Schuurmans, Quoc Le, Ed~Chi, and Denny Zhou.
\newblock Self-consistency improves chain of thought reasoning in language
  models.
\newblock \emph{arXiv preprint arXiv:2203.11171}, 2022.

\bibitem[Wei et~al.(2022)Wei, Wang, Schuurmans, Bosma, Chi, Le, and
  Zhou]{wei2022chain}
Jason Wei, Xuezhi Wang, Dale Schuurmans, Maarten Bosma, Ed~Chi, Quoc Le, and
  Denny Zhou.
\newblock Chain of thought prompting elicits reasoning in large language
  models.
\newblock \emph{arXiv preprint arXiv:2201.11903}, 2022.

\bibitem[Wu et~al.(2022)Wu, Xiao, Sun, Zhang, Ma, and He]{wu2022survey}
Xingjiao Wu, Luwei Xiao, Yixuan Sun, Junhang Zhang, Tianlong Ma, and Liang He.
\newblock A survey of human-in-the-loop for machine learning.
\newblock \emph{Future Generation Computer Systems}, 2022.

\bibitem[Yang et~al.(2015)Yang, Ma, Nie, Chang, and Hauptmann]{yang2015multi}
Yi~Yang, Zhigang Ma, Feiping Nie, Xiaojun Chang, and Alexander~G Hauptmann.
\newblock Multi-class active learning by uncertainty sampling with diversity
  maximization.
\newblock \emph{International Journal of Computer Vision}, 113\penalty0
  (2):\penalty0 113--127, 2015.

\bibitem[Zanzotto(2019)]{zanzotto2019human}
Fabio~Massimo Zanzotto.
\newblock Human-in-the-loop artificial intelligence.
\newblock \emph{Journal of Artificial Intelligence Research}, 64:\penalty0
  243--252, 2019.

\bibitem[Zhang et~al.(2022)Zhang, Roller, Goyal, Artetxe, Chen, Chen, Dewan,
  Diab, Li, Lin, et~al.]{zhang2022opt}
Susan Zhang, Stephen Roller, Naman Goyal, Mikel Artetxe, Moya Chen, Shuohui
  Chen, Christopher Dewan, Mona Diab, Xian Li, Xi~Victoria Lin, et~al.
\newblock Opt: Open pre-trained transformer language models.
\newblock \emph{arXiv preprint arXiv:2205.01068}, 2022.

\end{thebibliography}

\maketitle
\appendix

\section{Experiments for Filtering Stage}
\label{appendix:experiments_for_filtering_stage}

After the first stage, the LLM samples plausible rationales for a problem that arrive at different answers.
Just like humans, LLMs may make countless and various mistakes, but there are only a limited number of correct rationales for the right result.
If most of the sampled rationales cannot make agreements, with a high probability this sample is wrongly predicted.
To empirically prove that, we conduct quantitative experiments and discover that incorrectly predicted samples tend to have greater diversity in their final answer when solving difficult reasoning problems.

Specifically, the LLM is prompted with a set of manually written CoT exemplars following ~\citet{wei2022chain} in the first stage. Then, we sample a set of $5$ candidate outputs from the LLM's decoder to generate a set of rationales. Based on the sampled rationales, we divide the samples into two parts: \textbf{Part} 1 has all sampled rationales pointing to the same final answer (\ie, the Diversity Entropy score as Sec.~\ref{subsection:filtering} of such samples should be equal to $0$); \textbf{Part 2} has sampled rationales pointing to different final answers, which is the part outside the first part of samples (\ie, the Diversity Entropy score as Sec.~\ref{subsection:filtering} of such samples should be greater than $0$). Next, we calculate the accuracy of \textbf{Part 1} and \textbf{Part 2} for each dataset separately. We use the first answer of each sample as the result of CoT- Prompting and use all five answers to calculate the Diversity Entropy score. The results are shown in Tab.~\ref{table:expriment_for_filtering_stage_first}, Tab.~
\ref{table:expriment_for_filtering_stage_second}, Tab.~\ref{table:expriment_for_filtering_stage_third} and Tab.~\ref{table:expriment_for_filtering_stage_fourth}.
The accuracy of \textbf{Part 1} is generally larger than \textbf{Part 2}. It demonstrates the superiority of Diversity Entropy and experimentally confirms the intuition that incorrectly predicted samples tend to have greater diversity in their final answer when solving difficult reasoning problems.

\begin{table*}[ht]
\centering
\small
\setlength\tabcolsep{3pt}
    \centering
    \begin{tabular}{llcccccclclc}
    \toprule
    \multirow{3}{*}{Method} & \multirow{3}{*}{Part}&\multicolumn{9}{c}{\textit{Arithmetic Reasoning}} \\
    \cmidrule(r){3-11}
     & & \multicolumn{3}{c}{AddSub} & \multicolumn{3}{c}{MultiArith} & \multicolumn{3}{c}{SingleEq}  \\
    \cmidrule(r){3-5}
    \cmidrule(r){6-8}
    \cmidrule(r){9-11}
    & & Num. & Ratio & Acc. & Num. & Ratio & Acc. & Num. & Ratio & Acc. \\
    \midrule
    \multirow{3}{*}{CoT-Prompting} & Part 1 & 245 & 62.03$\%$ & 97.55 & 299 & 49.83$\%$ & 100.00 & 369 & 72.64$\%$ & 97.83 \\
                                  & Part 2 & 150 & 37.97$\%$ & 53.33 & 301 & 50.17$\%$ & 82.39 & 139 & 27.36$\%$ & 51.08 \\
                                    & Part 1\&2 & 395 & 100.00$\%$ & 82.78  & 600 & 100.00$\%$ & 93.00 & 508 & 100.00$\%$ & 85.04 \\
    \midrule
    \multirow{3}{*}{Self-Consistency} & Part 1 & 245 & 62.03$\%$ & 97.55 & 299 & 49.83$\%$ & 100.00 & 369 & 72.64$\%$ & 97.83 \\
                                     & Part 2 & 150 & 37.97$\%$ & 71.33 & 301 & 50.17$\%$ & 87.38  & 139 & 27.36$\%$ & 66.19 \\
                                       & Part 1\&2 & 395 & 100.00$\%$ & 90.63 & 600 & 100.00$\%$ & 94.17 & 508 & 100.00$\%$ & 89.17\\
    \bottomrule
    \end{tabular}
    \caption{Analysis for Diversity Entropy in Filtering Stage (I). The accuracy of \textbf{Part 1} is generally larger than \textbf{Part 2}. The result demonstrates the superiority of Diversity Entropy and experimentally confirms the intuition that incorrectly predicted samples tend to have greater diversity in their final answer when solving difficult reasoning problems. For each task, we report the median scores among 5 runs.}
    \label{table:expriment_for_filtering_stage_first}
\end{table*}

\begin{table*}[ht]
\centering
\small
\setlength\tabcolsep{3pt}
    \centering
    \begin{tabular}{llcccccccccc}
    \toprule
    \multirow{3}{*}{Method} & \multirow{3}{*}{Part}&\multicolumn{9}{c}{\textit{Arithmetic Reasoning}} \\
    \cmidrule(r){3-11}
     & & \multicolumn{3}{c}{SingleOp} & \multicolumn{3}{c}{ASDiv} & \multicolumn{3}{c}{AQuA}  \\
    \cmidrule(r){3-5}
    \cmidrule(r){6-8}
    \cmidrule(r){9-11}
    & & Num. & Ratio & Acc. & Num. & Ratio & Acc. & Num. & Ratio & Acc. \\
    \midrule
    \multirow{3}{*}{CoT-Prompting} & Part 1 & 423 & 75.27$\%$ & 98.35 & 1122 & 53.53$\%$ & 96.88 & 48 & 18.90$\%$ & 52.08 \\
                                  & Part 2 & 139 & 24.73$\%$ & 58.99 & 974 & 46.47$\%$ & 42.51 & 206 & 81.10$\%$ & 37.38 \\
                                    & Part 1\&2 & 562 & 100.00$\%$ & 94.84 & 2096 & 100.00$\%$ & 73.19 & 254 & 100.00$\%$ & 40.55 \\
    \midrule
    \multirow{3}{*}{Self-Consistency} & Part 1 & 423 & 75.27$\%$ & 98.35 & 1122 & 53.53$\%$ & 96.88 & 48 & 18.90$\%$ & 52.08 \\
                                     & Part 2 & 139 & 24.73$\%$ & 70.50 & 974 & 46.47$\%$ & 52.78 & 206 & 81.10$\%$ & 32.04 \\
                                       & Part 1\&2 & 562 & 100.00$\%$ & 95.73 & 2096 & 100.00$\%$ & 77.72 & 254 & 100.00$\%$ & 38.19 \\
    \bottomrule
    \end{tabular}
    \caption{Analysis for Diversity Entropy in Filtering Stage (II). The accuracy of \textbf{Part 1} is generally larger than \textbf{Part 2}. The result demonstrates the superiority of Diversity Entropy and experimentally confirms the intuition that incorrectly predicted samples tend to have greater diversity in their final answer when solving difficult reasoning problems. For each task, we report the median scores among 5 runs.}
    \label{table:expriment_for_filtering_stage_second}
\end{table*}

\begin{table*}[ht]
\centering
\small
\setlength\tabcolsep{3pt}
    \centering
    \begin{tabular}{llcccccccccc}
    \toprule
    \multirow{3}{*}{Method} & \multirow{3}{*}{Part} & \multicolumn{6}{c}{\textit{Arithmetic Reasoning}} & \multicolumn{3}{c}{\textit{Commonsense Reasoning}} \\
    \cmidrule(r){3-8}
    \cmidrule(r){9-11}
     & & \multicolumn{3}{c}{SVAMP} & \multicolumn{3}{c}{GSM8K} & \multicolumn{3}{c}{CSQA}  \\
    \cmidrule(r){3-5}
    \cmidrule(r){6-8}
    \cmidrule(r){9-11}
    & & Num. & Ratio & Acc. & Num. & Ratio & Acc. & Num. & Ratio & Acc. \\
    \midrule
    \multirow{3}{*}{CoT-Prompting} & Part 1 & 438 & 43.80$\%$ & 92.92 & 256 & 19.41$\%$ & 93.36 & 792 & 64.86$\%$ & 85.98 \\
                                  & Part 2 & 562 & 56.20$\%$ & 47.86 & 1063 & 80.59$\%$ & 47.70 & 429 & 35.14$\%$ & 47.09 \\
                                    & Part 1\&2 & 1000 & 100.00$\%$ & 68.00 & 1319 & 100.00$\%$ & 56.48 & 1221 & 100.00$\%$ & 72.32 \\
    \midrule
    \multirow{3}{*}{Self-Consistency} & Part 1 & 438 & 43.80$\%$ & 92.92 & 256 & 19.41$\%$ & 93.36 & 792 & 64.86$\%$ & 85.98 \\
                                     & Part 2 & 562 & 56.20$\%$ & 62.46 & 1063 & 80.59$\%$ & 50.71 & 429 & 35.14$\%$ & 57.81 \\
                                       & Part 1\&2 & 1000 & 100.00$\%$ & 75.70 & 1319 & 100.00$\%$ & 58.85 & 1221 & 100.00$\%$ & 76.09 \\
    \bottomrule
    \end{tabular}
    \caption{Analysis for Diversity Entropy in Filtering Stage (III). The accuracy of \textbf{Part 1} is generally larger than \textbf{Part 2}. The result demonstrates the superiority of Diversity Entropy and experimentally confirms the intuition that incorrectly predicted samples tend to have greater diversity in their final answer when solving difficult reasoning problems. For each task, we report the median scores among 5 runs.}
    \label{table:expriment_for_filtering_stage_third}
\end{table*}

\begin{table*}[ht]
\centering
\small
\setlength\tabcolsep{3pt}
    \centering
    \begin{tabular}{llcccccccccc}
    \toprule
    \multirow{3}{*}{Method} & \multirow{3}{*}{Part} & \multicolumn{3}{c}{\textit{Commonsense Reasoning}} & \multicolumn{6}{c}{\textit{Symbolic Reasoning}} \\
    \cmidrule(r){3-8}
    \cmidrule(r){9-11}
     & & \multicolumn{3}{c}{StrategyQA} & \multicolumn{3}{c}{Letter (4)} & \multicolumn{3}{c}{Coinflip (4)}  \\
    \cmidrule(r){3-5}
    \cmidrule(r){6-8}
    \cmidrule(r){9-11}
    & & Num. & Ratio & Acc. & Num. & Ratio & Acc. & Num. & Ratio & Acc. \\
    \midrule
    \multirow{3}{*}{CoT-Prompting} & Part 1 & 1502 & 65.88$\%$ & 66.31 & 175 & 35.00$\%$ & 72.00 & 384 & 38.40$\%$ & 98.70 \\
                                  & Part 2 & 778 & 34.12$\%$ & 48.59 & 325 & 65.00$\%$ & 36.31 & 616 & 61.60$\%$ & 69.48 \\
                                    & Part 1\&2 & 2280 & 100.00$\%$ & 60.13 & 500 & 100.00$\%$ & 49.20 & 1000 & 100.00$\%$ & 81.40 \\
    \midrule
    \multirow{3}{*}{Self-Consistency} & Part 1 & 1502 & 65.88$\%$ & 66.31 & 175 & 35.00$\%$ & 72.00 & 384 & 38.40$\%$ & 98.70 \\
                                     & Part 2 & 778 & 34.12$\%$ & 52.57 & 325 & 65.00$\%$ & 44.62 & 616 & 61.60$\%$ & 89.61 \\
                                       & Part 1\&2 & 2280 & 100.00$\%$ & 61.40 & 500 & 100.00$\%$ & 54.40 & 1000 & 100.00$\%$ & 93.20 \\
    \bottomrule
    \end{tabular}
    \caption{Analysis for Diversity Entropy in Filtering Stage (IV). The accuracy of \textbf{Part 1} is generally larger than \textbf{Part 2}. The result demonstrates the superiority of Diversity Entropy and experimentally confirms the intuition that incorrectly predicted samples tend to have greater diversity in their final answer when solving difficult reasoning problems. For each task, we report the median scores among 5 runs.}
    \label{table:expriment_for_filtering_stage_fourth}
\end{table*}

\section{Experiments for Correction Stage}
\label{appendix:experiments_for_correction_stage}

\subsection{Incorrect Rationale Could Output the Correct Final Answer after Manually Correcting the Erroneous Rationale.}

An incorrect rationale could output the correct final answer after correcting the erroneous rationale. To empirically prove this, we conduct quantitative experiments for twelve datasets and discover that in general most of the errors of CoT indeed are caused by incorrect rationales. After correcting these incorrect rationales, the final answers turn out to be correct.

Specifically, we explored the limits of the CoT-based methods (namely CoT-Prompting, Self-Consistency, and MCS) when humans correct rationales while disregarding cost. Humans were instructed to thoroughly check all samples and ensure the correctness of all rationales. Tables \ref{tabble:results_manual_correction_arithmetic_upperbound} and \ref{tabble:results_manual_correction_commonsense_symbolic__upperbound} present the results, where the upper bound of CoT-Prompting is denoted as CoT-Upperbound and the upper bound of Self-Consistency is denoted as SC-Upperbound. Self Consistency and MCS+Self Consistency have the same upper bound in extreme cases (\ie, the threshold of Diversity Entropy score is set to $100\%$) while CoT-Upperbound and MCS have the same upper bound in extreme cases (\ie, the threshold of Diversity Entropy score is set to $100\%$). The experimental results demonstrate that the upper bounds are quite high, indicating that an incorrect rationale could produce the correct final answer after correcting the errors. To note, this limitation represents only the upper bounds of our method, and its practical implementation would require significant time and resources.

\begin{table*}[htbp]
\centering
\small
\setlength\tabcolsep{3pt}
    \centering
        \begin{tabular}{ccccccccc}
    \toprule
    \multirow{2}{*}{Model} & \multicolumn{8}{c}{\textit{Arithmetic Reasoning}} \\
    \cmidrule(r){2-9}
     & AddSub & MultiArith & SingleEq & SingleOp & ASDiv & AQuA & SVAMP & GSM8K\\
    \midrule
    CoT-Prompting & 82.78 & 93.00 & 85.04 & 94.84 & 73.19 & 40.55 & 68.00 & 56.48 \\
    CoT-Upperbound & 97.72 & 96.33 & 94.09 & 96.80 & 75.62 & 47.64 & 77.50 & 63.76 \\
    \midrule
    Self-Consistency & 90.63 & 94.17 & 89.17 & 95.73 & 77.72 & 38.19 & 75.70 & 58.85 \\
    SC-Upperbound & 98.48 & 96.33 & 95.67 & 98.93 & 81.58 & 44.49 & 82.00 & 64.67 \\
    \bottomrule
    \end{tabular}
    \caption{Upperbound Analysis of CoT-Prompting, Self-Consistency and \model (I). The experimental results demonstrate that the upper bounds are quite high, indicating that an incorrect rationale could produce the correct final answer after correcting the errors. To note, this limitation represents only the upper bounds of our method, and its practical implementation would require significant time and resources. For each task, we report the median scores among 5 runs.}
    \label{tabble:results_manual_correction_arithmetic_upperbound}
\end{table*}

\begin{table}[htbp]
\centering
\small
\setlength\tabcolsep{3pt}
    \centering
    \begin{tabular}{ccccc}
    \toprule
    \multirow{2}{*}{Model} & \multicolumn{2}{c}{\textit{Commonsense}} & \multicolumn{2}{c}{\textit{Symbolic}} \\
    \cmidrule(r){2-3}
    \cmidrule(r){4-5}
     & CSQA & StraQA & Letter & Coinflip\\
    \midrule
    CoT-Prompting & 72.32 & 60.13 & 49.20 & 81.40\\
    CoT-Upperbound & 74.61 & 60.88 & 93.80 & 81.40 \\
    \midrule
    Self-Consistency & 76.09 & 61.40 & 54.40 & 93.20\\
    SC-Upperbound & 77.97 & 62.23 & 96.00 & 93.20 \\
    \bottomrule
    \end{tabular}
    \caption{Upperbound Analysis of CoT-Prompting, Self-Consistency and \model (II). The experimental results demonstrate that the upper bounds are quite high, indicating that an incorrect rationale could produce the correct final answer after correcting the errors. To note, this limitation represents only the upper bounds of our method, and its practical implementation would require significant time and resources. For each task, we report the median scores among 5 runs.}
    \label{tabble:results_manual_correction_commonsense_symbolic__upperbound}
\end{table}

\subsection{Correcting Erroneous Sub-logic Indeed Solves the Majority of Erroneous Rationale.}
\label{appendix:error_case_study}

Correcting erroneous sub-logic indeed solves the majority of erroneous rationale. We conduct the analytical experiment across multiple tasks in Sec.~\ref{subsection:error_case_study} and provide the evidence. 

We conduct experiments on twelve datasets to check whether correcting sub-logics solves the majority of incorrect rationales. Each task is represented by a pie chart. For each task, we conduct the error analysis for CoT prompting and analyze the error types of rationales. We divided the error types into four categories: errors that are able to be corrected by the ``modifying'' operation, the ``adding'' operation, the ``deleting'' operation, and the rest of the errors that are unable to be manually corrected. The percentage of each type across datasets is shown in Fig.~\ref{fig:error_case_study}.

Sec.~\ref{subsection:error_case_study} presents experiments in Fig.~\ref{fig:error_case_study} on twelve datasets to check whether correcting sub-logics solves the majority of erroneous rationales. Figure~\ref{fig:error_case_study} illustrates the error analysis of the CoT Prompting across twelve tasks.
We list the detailed numbers of the error analysis in Tab.~\ref{table:error_case_study_first} and Tab.~\ref{table:error_case_study_second}. Results show that correcting erroneous sub-logic indeed solves the majority of erroneous rationale (\ie, each erroneous rationale indeed can be corrected by only editing a single erroneous sub-logic).

\begin{table*}[ht]
\centering
\small
\setlength\tabcolsep{3pt}
    \centering
    \begin{tabular}{lcccccccccccc}
    \toprule
    \multirow{3}{*}{Operation} & \multicolumn{12}{c}{\textit{Arithmetic Reasoning}} \\
    \cmidrule(r){2-13}
     & \multicolumn{2}{c}{AddSub} & \multicolumn{2}{c}{MultiArith} & \multicolumn{2}{c}{SingleEq} & \multicolumn{2}{c}{SingleOp} & \multicolumn{2}{c}{ASDiv} & \multicolumn{2}{c}{AQuA}  \\
    \cmidrule(r){2-3}
    \cmidrule(r){4-5}
    \cmidrule(r){6-7}
    \cmidrule(r){8-9}
    \cmidrule(r){10-11}
    \cmidrule(r){12-13}
    & Num. & Ratio & Num. & Ratio & Num. & Ratio & Num. & Ratio & Num. & Ratio & Num. & Ratio  \\
    \midrule
    Modifying & 33 & 92$\%$ & 22 & 24$\%$ & 3 & 11$\%$ & 19 & 28$\%$ & 15 & 4$\%$ & 2 & 1$\%$ \\
     Adding & 0 &0$\%$ & 10 & 11$\%$ & 0 & 0$\%$ & 19 & 28$\%$ & 38 & 10$\%$ & 16 & 16$\%$ \\
     Deleting& 0 & 0$\%$ & 0 & 0$\%$ & 7 & 25$\%$ & 0 & 0$\%$ & 0 & 0$\%$ & 0 & 0$\%$ \\
    \midrule 
     Unable & 3 & 8$\%$ & 60 & 65$\%$ & 18 & 64$\%$ & 30 & 44$\%$ & 327 & 86$\%$ & 132 & 88$\%$ \\
    \bottomrule
    \end{tabular}
    \caption{Detailed numbers of the error analysis (I). The results are the detailed numbers of Fig.~\ref{fig:error_case_study}.}
    \label{table:error_case_study_first}
\end{table*}

\begin{table*}[ht]
\centering
\small
\setlength\tabcolsep{3pt}
    \centering
    \begin{tabular}{lcccccccccccc}
    \toprule
    \multirow{3}{*}{Operation} & \multicolumn{4}{c}{\textit{Arithmetic Reasoning}} & \multicolumn{4}{c}{\textit{Commonsense Reasoning}} & \multicolumn{4}{c}{\textit{Symbolic Reasoning}} \\
    \cmidrule(r){2-13}
     & \multicolumn{2}{c}{SVAMP} & \multicolumn{2}{c}{GSM8K} & \multicolumn{2}{c}{CSQA} & \multicolumn{2}{c}{StraQA} & \multicolumn{2}{c}{Letter (4)} & \multicolumn{2}{c}{Conflip (4)}  \\
    \cmidrule(r){2-3}
    \cmidrule(r){4-5}
    \cmidrule(r){6-7}
    \cmidrule(r){8-9}
    \cmidrule(r){10-11}
    \cmidrule(r){12-13}
    & Num. & Ratio & Num. & Ratio & Num. & Ratio & Num. & Ratio & Num. & Ratio & Num. & Ratio  \\
    \midrule
    Modifying & 41 & 13$\%$ & 54 & 10$\%$ & 28 & 8$\%$ & 39 & 36$\%$ & 223 & 88$\%$ & 0 & 0$\%$ \\
    Adding & 19 & 6$\%$ & 11 & 2$\%$ & 0 & 0$\%$ & 0 & 0$\%$ & 0 & 0$\%$ & 0 & 0$\%$ \\
    Deleting & 35 & 11$\%$ & 25 & 4$\%$ & 0 & 0$\%$ & 0 & 0$\%$ & 0 & 0$\%$ & 0 & 0$\%$ \\
    \midrule
     Unable & 225 & 70$\%$ & 478 & 84$\%$ & 310 & 92$\%$ & 69 & 64$\%$ & 30 & 12$\%$ & 186 & 100$\%$ \\
    \bottomrule
    \end{tabular}
    \caption{Detailed numbers of the error analysis (II). The results are the detailed numbers of Fig.~\ref{fig:error_case_study}.}
    \label{table:error_case_study_second}
\end{table*}

\subsection{Correcting Each Sub-logics Independently is Much Easier and More User-friendly than Correcting the Entire Rationale}

We conduct the human evaluation. The questionnaire survey shows that correcting each sub-logic independently (\ie, our approach) is much easier and more user-friendly than checking the entire rationale.
We present the time that humans need to check and correct the incorrect sub-logics compared to correcting the entire rationale as Tab.~\ref{tabble:results_manual_correction_arithmetic} and Tab.~\ref{tabble:results_manual_correction_commonsense_symbolic}.

The result presents the average time (seconds) needed for a human to check and correct the incorrect sub-logics compared to correcting the entire rationale for each sample.
The time humans need to check and correct the incorrect sub-logics is much less than the time needed to correct the entire rationale for each sample, proving that correcting each sub-logic independently is much easier and more user-friendly for humans than checking the entire rationale.

\begin{table*}[htbp]
\centering
\small
\setlength\tabcolsep{3pt}
    \centering
        \begin{tabular}{ccccccccc}
    \toprule
    \multirow{2}{*}{Human Operation} & \multicolumn{8}{c}{\textit{Arithmetic Reasoning}} \\
    \cmidrule(r){2-9}
     & AddSub & MultiArith & SingleEq & SingleOp & ASDiv & AQuA & SVAMP & GSM8K\\
    \midrule
    Correcting sub-logics & 21s & 24s & 30s & 14s & 26s & 62s & 16s & 45s \\
    Correcting entire rationale & 49s & 80s & 60s & 32s & 44s & 102s & 48s & 77s \\
    \bottomrule
    \end{tabular}
    \caption{Time (seconds) spent for correcting the incorrect sub-logics compared to correcting the entire rationale (I). The time humans need to check and correct the incorrect sub-logics is much less than the time needed to correct the entire rationale for each sample, proving that correcting each sub-logic independently is much easier and more user-friendly for humans than checking the entire rationale.}
    \label{tabble:results_manual_correction_arithmetic}
\end{table*}

\begin{table}[htbp]
\centering
\small
\setlength\tabcolsep{3pt}
    \centering
    \begin{tabular}{ccccc}
    \toprule
    \multirow{2}{*}{Human Operation} & \multicolumn{2}{c}{\textit{Commonsense}} & \multicolumn{2}{c}{\textit{Symbolic}} \\
    \cmidrule(r){2-3}
    \cmidrule(r){4-5}
     & CSQA & StraQA & Letter & Coinflip\\
    \midrule
    Correcting sub-logics & 14s & 24s & 18s & 36s \\
    Correcting entire rationale & 26s & 36s & 28s & 40s \\
    \bottomrule
    \end{tabular}
    \caption{Time (seconds) spent for correcting the incorrect sub-logics compared to correcting the entire rationale (II). The time humans need to check and correct the incorrect sub-logics is much less than the time needed to correct the entire rationale for each sample, proving that correcting each sub-logic independently is much easier and more user-friendly for humans than checking the entire rationale.}
    \label{tabble:results_manual_correction_commonsense_symbolic}
\end{table}

\section{Inference for CAMLOP}
\label{appendix:inference}

Given a model parameterized by $c, d$, and a fixed cost $y$, the model predicts the \textcolor{red}{optimal choice} $(x_1^{*}, x_2^{*})$ with the highest utility, which is desired by the company strategic decision-makers.
Note an important feature of this optimal choice: at this data point (namely, \textcolor{red}{optimal choice} point) the indifference curve is tangent to $p_1 x_1 + p_2 x_2 = y$. According to this feature, the inference is to get $(x_1^{*}, x_2^{*})$ that satisfied the following equation:
\begin{equation}
\begin{split}
u'(x_1^{*}, x_2^{*})=-\frac{p_1}{p_2}
\end{split}
\end{equation}
which will derive the optimal choice $(x_1^{*}, x_2^{*})$:
\begin{equation}
\begin{split}
x_1^{*} = \frac{c}{c+d} \frac{m}{p_1}, x_2^{*} = \frac{d}{c+d} \frac{m}{p_2}
\end{split}
\end{equation}



\section{Learning for CAMLOP}
\label{appendix:learning}

We have seen how to make the best decision based on the inference of CAMLOP. But in real life we have to work the other way around: we observe some historical cost and utility datapoints, but our problem is to estimate what kind of utility function is induced from the observations.

Concretely, suppose that we observe a number of industries making choices between LLMs and human workers based on their considerations of commute times, money costs, accuracy, \etc. There exists an analytic solution of $c,d$ obtained by statistical techniques that best fit the observed data points. In this way, the historical datapoints give a way to estimate the utility function.
More specifically, we use regression analysis to find the utility function that best describes the relation between $x$ and utility. Mean square error is typically employed as the loss function for learning the utility function. The loss function is defined on $J$ training datapoints $X=\{(x_1^{(1)},x_2^{(1)}),(x_1^{(2)},x_2^{(2)}),..., (x_1^{(J)},x_2^{(J)})\}$:
\begin{equation}\label{obj}
\begin{split}
L(c,d) =\frac{1}{J} \mathop{\sum_{i = 1}^{J}} \log u(x_1^{(i)},x_2^{(i)};c,d)
\end{split}
\end{equation}
where the model parameters are $c,d$. A normal equation or gradient descent can be used to optimize this loss function and obtain the final $c,d$.

\section{Experiment Details}
\label{appendix:experiment_details}

\begin{table*}[htbp]

    \centering
    \small
    \begin{tabular}{p{13.5cm}}
        \toprule
        \textbf{Q:} There are 15 trees in the grove. Grove workers will plant trees in the grove today. After they are done, there will be 21 trees. How many trees did the grove workers plant today? \\
        \textbf{A:} We start with 15 trees. Later we have 21 trees. The difference must be the number of trees they planted. So, they must have planted 21 - 15 = 6 trees. The answer is 6. \\
        \textbf{Q:} If there are 3 cars in the parking lot and 2 more cars arrive, how many cars are in the parking lot? \\
        \textbf{A:} There are 3 cars in the parking lot already. 2 more arrive. Now there are 3 + 2 = 5 cars. The answer is 5. \\
        \textbf{Q:} Leah had 32 chocolates and her sister had 42. If they ate 35, how many pieces do they have left in total? \\
        \textbf{A:} Leah had 32 chocolates and Leah's sister had 42. That means there were originally 32 + 42 = 74 chocolates. 35 have been eaten. So in total they still have 74 - 35 = 39 chocolates. The answer is 39. \\
        \textbf{Q:} Jason had 20 lollipops. He gave Denny some lollipops. Now Jason has 12 lollipops. How many lollipops did Jason give to Denny? \\
        \textbf{A:} Jason had 20 lollipops. Since he only has 12 now, he must have given the rest to Denny. The number of lollipops he has given to Denny must have been 20 - 12 = 8 lollipops. The answer is 8. \\
        \textbf{Q:} Shawn has five toys. For Christmas, he got two toys each from his mom and dad. How many toys does he have now? \\
        \textbf{A:} He has 5 toys. He got 2 from mom, so after that he has 5 + 2 = 7 toys. Then he got 2 more from dad, so in total he has 7 + 2 = 9 toys. The answer is 9. \\
        \textbf{Q:} There were nine computers in the server room. Five more computers were installed each day, from monday to thursday. How many computers are now in the server room? \\
        \textbf{A:} There are 4 days from monday to thursday. 5 computers were added each day. That means in total 4 * 5 = 20 computers were added. There were 9 computers in the beginning, so now there are 9 + 20 = 29 computers. The answer is 29. \\
        \textbf{Q:} Michael had 58 golf balls. On tuesday, he lost 23 golf balls. On wednesday, he lost 2 more. How many golf balls did he have at the end of wednesday? \\
        \textbf{A:} Michael initially had 58 balls. He lost 23 on Tuesday, so after that he has 58 - 23 = 35 balls. On Wednesday he lost 2 more so now he has 35 - 2 = 33 balls. The answer is 33.\ \\
        \textbf{Q:} Olivia has \$23. She bought five bagels for \$3 each. How much money does she have left? \\
        \textbf{A:} She bought 5 bagels for \$3 each. This means she spent 5 * \$3 = \$15 on the bagels. She had \$23 in beginning, so now she has \$23 - \$15 = \$8. The answer is 8. \\
        \bottomrule
    \end{tabular}
        \caption{
    Few-shot exemplars for arithmetic reasoning tasks.
    }
    \label{tab:prompt-math}
\end{table*}

We choose GPT-3 because of its superior CoT reasoning performance, as reported in the work of \citet{wei2022chain} and \citet{wang2022self}. Due to the limited context window size (up to 4096 word-pieces for the GPT-3 series of models), we use an 8-shot setting for all datasets. Our experiments are based on access to the OpenAI GPT-3 API. We perform all experiments in the few-shot setting, without training or fine-tuning the LLM. For a fair comparison, we use the same prompts as in the work of \citet{wei2022chain}. For arithmetic reasoning tasks, we use the same set of 8 manually written exemplars.
For commonsense reasoning tasks, exemplars are randomly selected from the training set with manually written CoT prompts. 

We list the exact set of prompts used for all arithmetic reasoning tasks in Tab.~\ref{tab:prompt-math}, since there are multiple sets of prompts introduced in ~\citet{wei2022chain}. 
The prompts for CommonsenseQA and StrategyQA are the same as used in ~\citet{wei2022chain}.

\section{Diversity Metrics Over Diverse Reasoning Paths}
\label{appendix:diversity_metrics_over_diverse_reasoning_paths}

As described in Sec.~\ref{paragraph:diversity_metrics_over_diverse_reasoning_paths}, the majority vote method of calculating the answer probability over all sampled rationales can be regarded as taking an unnormalized unweighted sum. As described in \citet{wang2022self}, other methods of computing answer probability of $\mathbf{a}$ include the unnormalized weighted average, normalized weighted average, unnormalized weighted sum, and normalized weighted sum.
Tab.~\ref{table:aggregation} shows that unnormalized unweighted sum generally outperforms others. We use this setting in all experiments following ~\citet{wang2022self}.

In practice, the majority vote method of calculating the answer probability over all sampled rationales proposed at Eq.~\ref{equation:answer_probability} is the same as taking the unweighted sum over $\mathbf{a}_i$ (\ie, ${\textstyle \sum_{i=1}^{ | N |}} \mathbf{1}  (\mathbf{a}_i = \mathbf{a})$), where $|N|$ denotes the number of answers (\ie, the number of sampling times). As described in \citet{wang2022self}, another selection of computing answer probability of $\mathbf{a}$ over all sampled rationales is to use unnormalized probability $\mathbf{p}_{\mathbf{a}_i}$ of the language model generating $\mathbf{a}_i$ given the prompt of sample $\mathbf{s}$:

\begin{equation}
\mathbf{p}_{\mathbf{a}_i} = P(\mathbf{r}_i, \mathbf{a}_i \mid \mathbf{s})
\label{equation:unnormalized_probability}
\end{equation}

Then we use all unnormalized probability $\mathbf{p}_{\mathbf{a}_i}$ given by the language model's decoder to calculate the probability $\mathbf{p}_\mathbf{a}$ of the answer $\mathbf{a}$ for sample $\mathbf{s}$:

\begin{equation}
\mathbf{p}_\mathbf{a} = \frac{{\textstyle \sum_{i=1}^{\left | N \right |}} \mathbf{1}  (\mathbf{a}_i = \mathbf{a}) \mathbf{p}_{\mathbf{a}_i} }{\left | N \right | }
\label{equation:unnormalized_sum}
\end{equation}

where $|N|$ denotes the number of rationales decoded for the sample $\mathbf{s}$. The result of using the calculation output of Eq.~\ref{equation:unnormalized_sum} as the probability of answer $\mathbf{a}$ is shown in Tab.~\ref{table:aggregation} as \textbf{Unnormalized Weighted Sum }. Apart from computing $\mathbf{p}_\mathbf{a}  $ by taking the unnormalized probability of the language model generating $(\mathbf{r}_i, \mathbf{a}_i)$ given $\mathbf{s}$, we can normalize the output probability for $(\mathbf{r}_i, \mathbf{a}_i)$ by the output length of $\mathbf{r}_i$ \citep{brown2020language}: 

\begin{equation}
\mathbf{p}_{\mathbf{a}_i} = \exp^{\frac{1}{K}\sum_{k=1}^K {\log p_{t_k}}}
\label{equation:normalized_probability}
\end{equation}

where $p_{t_k}$ is the log probability of generating the $k$-th token $t_k$ in $(\mathbf{r}_i, \mathbf{a}_i)$ conditioned on the previous tokens, and $K$ is the total number of tokens in $(\mathbf{r}_i, \mathbf{a}_i)$:

\begin{equation}
p_{t_k} = P(t_k \mid \mathbf{s}, t_1, \ldots, t_{k-1})
\label{equation:gentate_token}
\end{equation}

The result of using the calculation output of Eq.~\ref{equation:normalized_probability} as the normalized probability $\mathbf{p}_i^\mathbf{a}$ of the language model generating $\mathbf{a}_i$ given prompt of sample $\mathbf{s}$ is shown in Tab.~\ref{table:aggregation} as \textbf{Normalized Weighted Sum}.


In addition, in Tab.~\ref{table:aggregation} we also report the results by taking a weighted average, which means calculating a score for each $\mathbf{a}$ of its weighted sum divided by $\sum\nolimits_{i=1}^{|N|} \mathbf{1} (\mathbf{a}_i = \mathbf{a})$.

Tab.~\ref{table:aggregation} shows that unnormalized unweighted sum generally outperforms others. We use this setting in all experiments following ~\citet{wang2022self}.

\section{Details of Balancing Cost and Utility}
\label{appendix:details_of_balancing_cost_and_utility}

In Sec~\ref{tab:utility-cost}, we conduct experiments on the SingleEq dataset to quantitatively calculate cost and utility for \cail. The trends on other datasets are consistent with SingleEq dataset. We randomly selected one dataset as an example to demonstrate the superiority of MCS in balancing cost and utility.

For the cost, we consider money and time.
We set the price of the LLM as $\mathbf{p}_{llm}$ and the time cost as $\mathbf{t}_{llm}$. Since we use GPT-3, the price $\mathbf{p}_{llm}$ for a single math problem (decoding once) is $\$0.08$ on average, and the time cost $\mathbf{t}_{llm}$ is $0.8$ second based on empirical results
\footnote{The pricing of \texttt{text-davinci-002} is $\$0.02$ per $1000$ tokens, which can be found at  \url{https://openai.com/pricing}. We set $\mathbf{p}_{llm}$ to be $\$0.08$ because an input sample for few-shot CoT contains about $4000$ tokens on average when decoding only once. Note that we only calculated the time for the main part (\ie, the decoding) and ignored other parts that were fast enough to be ignored compared to the API calls.}.
The price of solving a single math problem with only human labor is $\mathbf{p}_{human}$ and the time cost is $\mathbf{t}_{human}$.  We set $\mathbf{p}_{human}$ to be $\$0.125$ and $\mathbf{t}_{human}$ to be $60$ seconds based on our empirical results.
\footnote{Minimum hourly wage in the United States is $\$7.5$, which can be found at \url{https://www.worker.gov/pay-for-hours-worked/}.
Solving a problem requires $60$ seconds on average. Therefore, the price and time cost required to complete a problem are $\$0.125$ and $60$ seconds, respectively.}
The price of human labor for \model to correct a single math problem $\mathbf{p}_{\model}$ is $\$0.0625$ and the time cost $\mathbf{t}_{\model}$ is $30$ seconds based on empirical results. Note the time required to inspect and correct is less than the time needed to fully solve the entire problem, therefore $\mathbf{t}_{\model} < \mathbf{t}_{human}$.

For the utility, we consider user satisfaction as the comprehensive score. We ask five users to write down their satisfaction levels and calculate the average. The human ratings are collected via Amazon Turk. In addition to the effective data collected from $5$ users for each evaluation method, data from several users were excluded due to failures in the attention verification. The hourly salary is $\$$10 per hour and per user. We randomly select a set of examples and the satisfaction level is rated from $1$ to $5$, with $1$ as the worst satisfaction and $5$ as the most user-friendly and best satisfaction. The human rating scores are then averaged. 

\begin{table}[htbp]
\centering
\small
\setlength\tabcolsep{3pt}
    \centering
        \begin{tabular}{lcccc}
    \toprule
    Plans & Time & Money & Acc. & Utility(User Satis.) \\
    \midrule
    Human & $60$s & $\$0.125$ & 93.20 & 86.40 \\
    \midrule
    CoT Prompting & $0.8$s & $\$0.080$ & 85.04 & 81.60 \\
    Self-Consistency ($\mathbf{N}_{self} = 10$) & $8$s & $\$0.800$ & 92.49 & 85.80 \\
    \midrule
    \model ($\mathbf{N}_{\model} = 5$, $\mathbf{\alpha} = 20\%$) & $10.8$s & $\$0.4925$ & 91.00 & 84.20 \\
    \model + Self-consistency ($\mathbf{N}_{\model} = 5$, $\mathbf{\alpha} = 20\%$) & $10.8$s & $\$0.4925$ & 93.50 & 88.80 \\
    \midrule
    \model ($\mathbf{N}_{\model} = 5$, $\mathbf{\alpha} = 40\%$) & $16.8$s & $\$0.505$ & 92.51 & 85.60 \\
    \model + Self-consistency ($\mathbf{N}_{\model} = 5$, $\mathbf{\alpha} = 40\%$) & $16.8$s & $\$0.505$ & 94.09 & 90.80 \\ 
    \bottomrule
    \end{tabular}
    \caption{Analysis of cost and utility for SingleEq.
    \model + Self-consistency generally outperforms other methods with higher utility and acceptable cost. $\mathbf{N}_{\cdot}$: \# sampled rationale. $\mathbf{\alpha}$: DE threshold. Acc.: Accuracy. User Satis.: User Satisfaction.}
    \label{tab:utility-cost-appendix}
\end{table}


We experiment on candidate plans based on models from Sec.~\ref{sec:main_results} and Sec. \ref{section:additional_study} (Fig.~\ref{fig:threshold} and Fig.~\ref{fig:numbers}), and the results are shown in Tab.~\ref{tab:utility-cost-appendix}.
The calculation of time and money in Tab.~\ref{tab:utility-cost-appendix} is shown as below:
\begin{enumerate}[]
\item \textit{Human}: A plan that requires only human labor, which costs $\mathbf{p}_{human}$ and $\mathbf{t}_{human}$ seconds. So the time needed is $\mathbf{t}_{human} = 60$seconds, and the money needed is $\mathbf{p}_{human} = \$0.125$
\item \textit{CoT-prompting}: A naive CoT plan that only requires GPT-3 for decoding only once, which costs $\mathbf{p}_{llm}$ and $\mathbf{t}_{llm}$ seconds. So the money needed is $\mathbf{p}_{llm}=\$0.08$ and the time needed is $\mathbf{t}_{llm}=0.8$second.
\item \textit{Self-consistency ($\mathbf{N}_{self} = 10$)}: A Self-consistency plan that requires only LLMs to sample from the decoder $\mathbf{N}_{self}$ times, which will cost $ \mathbf{N}_{self} * \mathbf{p}_{llm}$ and $\mathbf{N}_{self} * \mathbf{t}_{llm} $ seconds. For $\mathbf{N}_{self} = 10$, the money needed is $\mathbf{N}_{self} * \mathbf{p}_{llm} = 10 * \$0.08 = \$0.8$, the time needed is $\mathbf{N}_{self} * \mathbf{t}_{llm} = 10 * 0.8 = 8 $seconds.
\item \textit{\model ($\mathbf{N}_{\model} = 5$, $\mathbf{\alpha} = 20\%$)}: \model samples from LLM decoder $\mathbf{N}_{\model}$ times and uses top $\mathbf{\alpha}$ as threshold, requiring $ (\mathbf{N}_{\model}+1) * \mathbf{p}_{llm} + \mathbf{\alpha} * \mathbf{p}_{\model}$ and $ (\mathbf{N}_{\model}+1) * \mathbf{t}_{llm} + \mathbf{\alpha} * \mathbf{t}_{\model}$ seconds. For $\mathbf{N}_{\model} = 5$, $\mathbf{\alpha} = 20\%$, the money needed is $ (\mathbf{N}_{\model}+1) * \mathbf{p}_{llm} + \mathbf{\alpha} * \mathbf{p}_{\model} = \$0.08 * 6 + 20\% * \$0.0625 = \$0.4925$, the time needed is $ (\mathbf{N}_{\model}+1) * \mathbf{t}_{llm} + \mathbf{\alpha} * \mathbf{t}_{\model} = 0.8 * 6s + 20\% * 30s = 10.8$ seconds.
\item \textit{\model + Self-consistency ($\mathbf{N}_{\model} = 5$, $\mathbf{\alpha} = 20\%$)}: A \model + Self-consistency ($\mathbf{N}_{\model} = 5$, $\mathbf{\alpha} = 20\%$) plan that requires to sample from the decoder $\mathbf{N}_{\model}$ times, which costs the same as the \model ($\mathbf{N}_{\model} = 5$, $\mathbf{\alpha} = 20\%$) plan.
\item \textit{\model ($\mathbf{N}_{\model} = 5$, $\mathbf{\alpha} = 40\%$)}: \model samples from LLM decoder $\mathbf{N}_{\model}$ times and uses top $\mathbf{\alpha}$ as threshold, requiring $ (\mathbf{N}_{\model}+1) * \mathbf{p}_{llm} + \mathbf{\alpha} * \mathbf{p}_{\model}$ and $ (\mathbf{N}_{\model}+1) * \mathbf{t}_{llm} + \mathbf{\alpha} * \mathbf{t}_{\model}$ seconds. For $\mathbf{N}_{\model} = 5$, $\mathbf{\alpha} = 40\%$, the money needed is $ (\mathbf{N}_{\model}+1) * \mathbf{p}_{llm} + \mathbf{\alpha} * \mathbf{p}_{\model} = \$0.08 * 6 + 40\% * \$0.0625 = \$0.505$, the time needed is $ (\mathbf{N}_{\model}+1) * \mathbf{t}_{llm} + \mathbf{\alpha} * \mathbf{t}_{\model} = 0.8 * 6s + 40\% * 30s = 16.8$ seconds.
\item \textit{\model + Self-consistency ($\mathbf{N}_{\model} = 5$, $\mathbf{\alpha} = 40\%$)}: A \model + Self-consistency ($\mathbf{N}_{\model} = 5$, $\mathbf{\alpha} = 40\%$) plan that requires to sample from the decoder $\mathbf{N}_{\model}$ times, which costs the same as the \model ($\mathbf{N}_{\model} = 5$, $\mathbf{\alpha} = 40\%$) plan.

\end{enumerate}

The results are shown in Tab.~\ref{tab:utility-cost-appendix}.
 The result shows that \model+Self-consistency generally outperforms other methods with higher utility (\ie, better user satisfaction) as well as an acceptable cost.

We performed regression analysis on user satisfaction based on LLM and Human and ultimately learned the utility function $\mathbf{u}(\mathbf{x}_{LLM}, \mathbf{x}_{Human}) = \mathbf{x}_{LLM}^{2.05}*(10 * \mathbf{x}_{Human})^{1.94}$, where $\mathbf{x}_{LLM}$ equals to $1$ when using LLM to decode one time, and $\mathbf{x}_{Human}$ equals to 10 when solving the problem with only human.

\section{Related Work}
\label{appendix:related_work}

\subsection{Human-In-the-Loop System}
The human-in-the-Loop system, aiming to achieve what neither humans nor machines can accomplish independently, is defined as a model requiring human interaction ~\citep{karwowski2006international}. When the machine cannot solve the problem, or when cost or security considerations require humans to participate, manual intervention is necessary ~\citep{wu2022survey, zanzotto2019human, mosqueira2023human}. Previous human-in-the-loop systems focus either on adding appropriate tags to data or providing feedback on cases with a certain confidence interval to the machines and thus retrain the model afterward with the labeled data or rewarded cases~\citep{wu2022survey,zanzotto2019human}. 
The human-in-the-loop system outperforms both standalone AI and humans working alone ~\citep{bien2018deep}.

Recently, LLM-based AI (Artificial Intelligence) systems are developing very quickly, and this trend is expected to expand to the majority of the workforce in the near future~\citep{ouyang2022training, zhang2022opt, sanh2021multitask}. However, these systems do not always provide satisfactory answers without human intervention, especially mathematical problems. Additionally, in domains such as criminal fact identification and charge predictions, inference should be reasonable and controlled by humans ~\citep{custers2022ai} while LLMs are not qualified.
Therefore, it is essential to develop a human-in-the-loop prompting-based system that is designed with the ability to collaborate with people. Such a system would make work more efficient and effective. Until recently, few researchers have systematically and quantitatively explored human-in-the-loop prompting-based systems.

Different from ChatGPT's RLHF (Reinforcement Learning from Human Feedback)~\footnote{\url{https://openai.com/blog/chatgpt}.}, we take the first step to use human feedback in an online way without access to parameters. Even though it's a preliminary step, this online method could benefit from further refinement and combination with RLHF in future research.

\subsection{In-context Learning}
Over the past decade, there have been significant advancements in Large Language Models (LLMs)~\citep{ouyang2022training, zhang2022opt, sanh2021multitask}. These developments have been further accelerated by the introduction of In-Context Learning (ICL)~\citep{kojima2022large}. Essentially, LLMs are capable of processing a few training examples and a test instance as its natural language instruction. It then directly decodes the output without requiring any updates to its parameters.
LLMs can perform diverse tasks effectively when provided with corresponding instructions ~\citep{ouyang2022training,srivastava2022beyond,wei2022chain}.
This presents an opportunity for humans to modify predicted outcomes through natural language instructions, which serve as a flexible and user-friendly interface.

\subsection{Chain-of-Thought Prompting}
Chain-of-Thought (CoT) prompting enables models to decompose multi-step problems into smaller steps. With CoT, LLMs can solve complex reasoning problems that cannot be solved with standard prompting methods ~\citep{wei2022chain,wang2022self}.
Despite its usefulness, CoT may be prone to errors, which can have a negative impact on the reasoning of the model. 
Fortunately, most mistakes can be easily interpreted. About half of these mistakes are related to incorrect calculations while the other half are mistakes from flawed reasoning where rationales lack the necessary knowledge ~\citep{Minerva}. To address this issue, we limit users to modifying, deleting, or adding a single sub-logic as a means of resolving both types of errors.
Additionally, we have found that most mistakes can be easily detected and corrected by humans through rationales. Against this background, CoT presents an opportunity for humans to efficiently modify predicted outcomes through sub-logics of rationales.

\end{document}